\pdfoutput=1

\documentclass[11pt]{article}

\usepackage[preprint]{acl}

\usepackage{times}
\usepackage{latexsym}

\usepackage[T1]{fontenc}

\usepackage[utf8]{inputenc}

\usepackage{microtype}

\usepackage{inconsolata}

\usepackage{graphicx}
\usepackage{amsmath}
\usepackage{amssymb}
\usepackage{multirow}
\usepackage{booktabs}
\usepackage{lipsum}
\usepackage{xcolor}
\definecolor{mygreen}{RGB}{1,113,0}
\usepackage{colortbl}
\definecolor{lightbluegray}{RGB}{235,245,255}
\definecolor{lightgreengray}{RGB}{235,245,235}
\usepackage{pifont}
\usepackage[utf8]{inputenc}
\usepackage{mdframed}
\usepackage{enumitem}
\usepackage[most]{tcolorbox}
\usepackage{algorithm,algpseudocode}
\usepackage{adjustbox}
\usepackage[utf8]{inputenc}
\usepackage[T1]{fontenc}   
\usepackage{hyperref}      
\usepackage{url}           
\usepackage{booktabs}      
\usepackage{amsfonts}      
\usepackage{nicefrac}      
\usepackage{microtype}     
\usepackage{wrapfig}
\usepackage{arydshln}
\usepackage{amsmath}
\usepackage{cuted}
\DeclareMathOperator*{\argmax}{arg\,max}

\title{T$^2$: An Adaptive Test-Time Scaling Strategy for \\Contextual Question Answering}

\author{
  \textbf{Zhengyi Zhao\textsuperscript{1}},
  \textbf{Shubo Zhang\textsuperscript{2}},
  \textbf{Zezhong Wang\textsuperscript{1}},
  \textbf{Huimin Wang\textsuperscript{3}},
  \textbf{Yutian Zhao\textsuperscript{3}},\\
  \textbf{Bin Liang\textsuperscript{1}},
  \textbf{Yefeng Zheng\textsuperscript{4}},
  \textbf{Binyang Li\textsuperscript{2}},
  \textbf{Kam-Fai Wong\textsuperscript{1}},
  \textbf{Xian Wu\textsuperscript{3,\thanks{Corresponding author.}}},
\\
\\
  \textsuperscript{1} The Chinese University of Hong Kong
  \textsuperscript{2} University of International Relations \\
  \textsuperscript{3} Jarvis Research Center, Tencent YouTu Lab 
  \textsuperscript{4} Westlake University
\\
  {
   \texttt{zyzhao@se.cuhk.edu.hk}
  }
}

\begin{document}
\maketitle
\begin{abstract}
Recent advances in Large Language Models (LLMs) have demonstrated remarkable performance in Contextual Question Answering (CQA). However, prior approaches typically employ elaborate reasoning strategies regardless of question complexity, leading to low adaptability. Recent efficient test-time scaling methods introduce budget constraints or early stop mechanisms to avoid overthinking for straightforward questions. But they add human bias to the reasoning process and fail to leverage models' inherent reasoning capabilities. To address these limitations, we present T$^2$: Think-to-Think, a novel framework that dynamically adapts reasoning depth based on question complexity. T$^2$ leverages the insight that if an LLM can effectively solve similar questions using specific reasoning strategies, it can apply the same strategy to the original question. This insight enables to adoption of concise reasoning for straightforward questions while maintaining detailed analysis for complex problems. T$^2$ works through four key steps: decomposing questions into structural elements, generating similar examples with candidate reasoning strategies, evaluating these strategies against multiple criteria, and applying the most appropriate strategy to the original question. Experimental evaluation across seven diverse CQA benchmarks demonstrates that T$^2$ not only achieves higher accuracy than baseline methods but also reduces computational overhead by up to 25.2\%. 
\end{abstract}

\section{Introduction}

Large language models (LLMs) have demonstrated impressive capabilities in Contextual Question Answering (CQA) tasks \cite{DBLP:conf/acl/TrivediBKS23,DBLP:conf/emnlp/PressZMSSL23}, but their reasoning approaches often lack adaptability to question complexity. Current CQA systems typically employ either direct answer generation or elaborate step-by-step reasoning for all questions, regardless of difficulty~\cite{wei2022chain,huang2024o1,min2024imitate}. This one-size-fits-all approach has an accuracy-vs-efficiency delimma. Directly generating answers for all questions will deteriorate the performance on difficult questions, which require multi-hop reasoning. Elaborated reasoning for all questions creates an efficiecy challenge: models frequently generate reasoning chains that are excessively verbose, containing redundant steps that do not contribute to finding the correct answer.

Existing analysis reveals that these redundant reasoning paths can unnecessarily extend the length of reasoning chains multiple times beyond what is required. Such as exploring multiple solution approaches when only one is needed \cite{ji2025first}, or verifying simple facts with elaborate explanations \cite{muennighoff2025s1}. For example, when asked ``What is the capital of France?'', models often generate lengthy discussions about France's history and geography before providing the straightforward answer ``Paris.'' This computational inefficiency is particularly concerning as model deployment costs continue to rise. Recent studies on reasoning efficiency~\cite{yang2025towards,zeng2025revisiting} confirm that blindly increasing reasoning chain length can actually harm performance on simpler tasks. Various attempts have been made to address this through adding a budget or stop mechanism to test-time scaling (TTS) methods~\cite{wei2022chain,huang2024o1} to stop thinking early, but these approaches introduce a human bias to the reasoning process~\cite{yuan2023scaling} and fail to leverage the model's inherent reasoning abilities.

Hence, the fundamental challenge is to develop a reasoning mechanism that can dynamically adjust its computational effort based on question complexity, which means providing concise reasoning for straightforward questions while maintaining detailed analysis for complex problems. Therefore, we present T$^2$, a think-to-think framework for efficient TTS strategy. T$^2$ leverages a key insight: if an LLM can effectively solve similar questions using specific reasoning strategies, it can apply comparable strategies to the original question. The process involves four key steps: (1) Decomposing the original question into its structural elements. For example, given the question:
\begin{quote}
    [Given Reference Documents]\\
    \textit{``Which is taller, the Eiffel Tower or the Empire State Building?''}
\end{quote}
T$^2$ would identify this as a comparative question involving measurement between two specific places as ``Which is [adj], [place 1] or [place 2]?''. (2) Creating a diverse set of similar example questions with the same question structure, each paired with supporting documents and potential reasoning strategies. Each reasoning strategy breaks down similar questions into simpler steps using fundamental reasoning skills (e.g., decomposing similar question ``Which is taller, Building A or Building B?'' into subquestions about individual heights connected by deductive reasoning for comparison). (3) Evaluating these reasoning strategies using multiple criteria to select the most appropriate strategy for the original question. (4) Applying the selected reasoning strategy to the original question while filtering irrelevant information.

By learning from similar examples, the model develops a more nuanced understanding of when detailed reasoning is necessary and when a more direct approach is sufficient. This allows T$^2$ to balance accuracy and efficiency without relying on pre-determined reasoning templates.

We evaluate T$^2$ across seven diverse CQA datasets ranging from simple factual queries to complex multi-hop reasoning tasks. Our results demonstrate that T$^2$ achieves superior accuracy (up to a 21.3\% increase) compared to other TTS approaches while reducing computational requirements by up to 25.2\%. These efficiency gains are particularly clear for simpler questions where redundant reasoning steps are eliminated. While for complex questions, T$^2$ maintains the reasoning depth required for accuracy without exploring unnecessary paths.

Our contributions include:
\begin{itemize}[leftmargin=*, itemsep=0pt,parsep=0pt,topsep=0pt,partopsep=0pt]
    \item We introduce T$^2$, a framework that enables language models to dynamically select appropriate reasoning strategies through similar examples, balancing efficiency and thoroughness based on question complexity.
    \item We develop a multi-criteria selection method that evaluates potential reasoning strategies based on coverage and uniqueness, ensuring the most suitable approach is applied to each question.
    \item We demonstrate through extensive experiments across diverse CQA benchmarks that our method reduces computational requirements by up to 25.2\% with superior accuracy.
\end{itemize}

\section{Related Work}

\paragraph{Contextual QA.} In addressing contextual QA, recent works have explored multi-round retrieval or reasoning approaches, including query rewriting for subsequent retrievals~\cite{DBLP:journals/corr/abs-2212-14024,DBLP:journals/corr/abs-2305-14283,DBLP:conf/emnlp/ShaoGSHDC23, DBLP:conf/emnlp/JiangXGSLDYCN23}, alternating between retrieval and reasoning steps~\cite{DBLP:conf/acl/TrivediBKS23}, and employing multi-round self-asking techniques~\cite{DBLP:conf/emnlp/PressZMSSL23}. They all rely on LLMs' reasoning abilities. 

\begin{figure*}[!t]
    \centering
    \includegraphics[width=\linewidth]{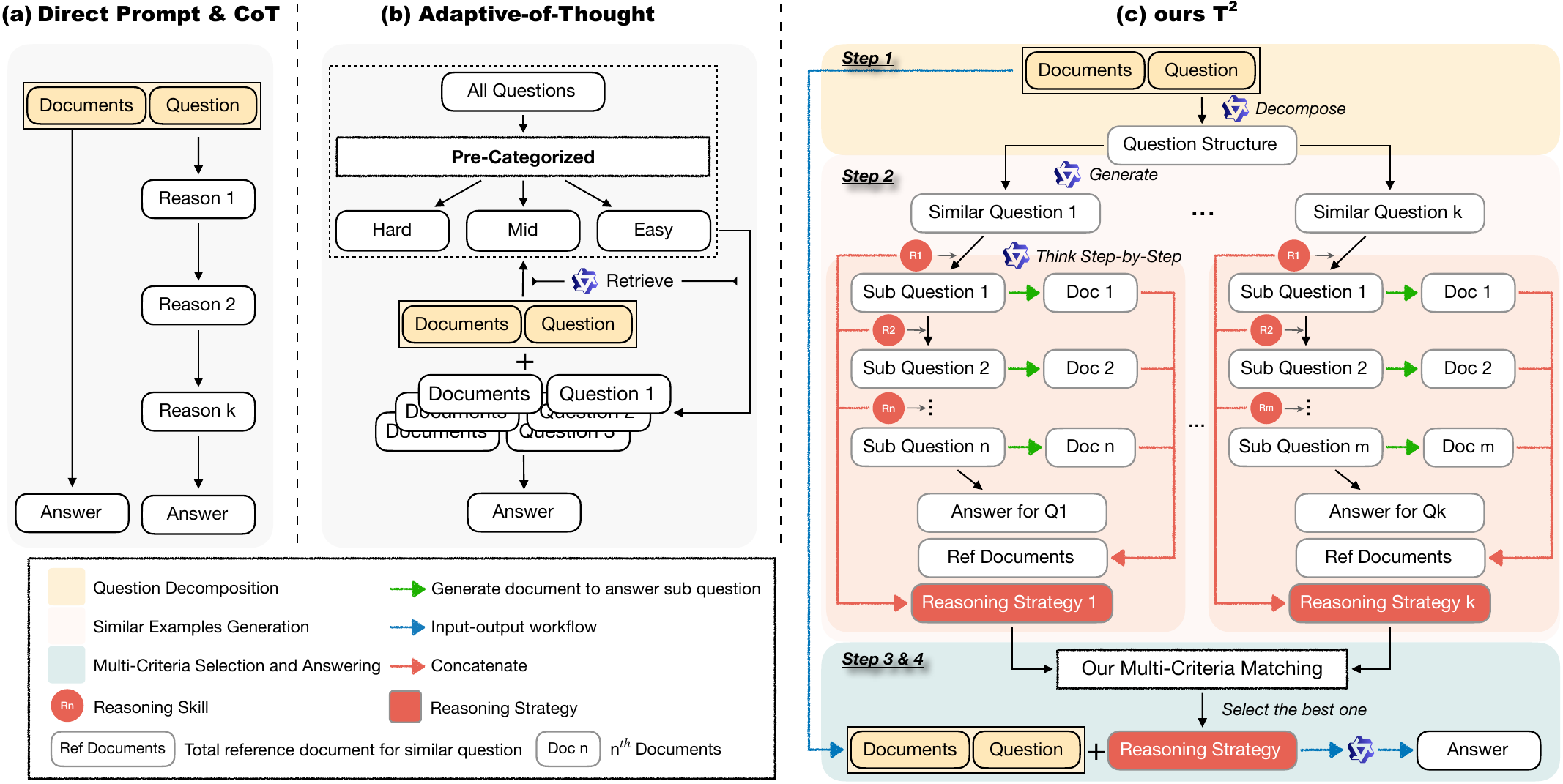}
    \caption{Overview of our T$^2$. \textbf{(a)} direct prompt or Chain-of-Thought (CoT), which adopts the same reasoning strategy regardless of question complexity. \textbf{(b)} Adaptive-of-Thought, which designs a question complexity evaluator to pre-categorize all questions, which might bring human bias in the evaluator design process. \textbf{(c)} our T$^2$. Instead of pre-categorizing questions into different complexity sets, T$^2$ generates multiple similar examples for different inputs adaptively and selects the best reasoning strategy for answering.}
    \label{fig:overview}
\end{figure*}

\paragraph{Test-Time Scaling.} Recent approaches to enhancing LLM reasoning capabilities focus on increasing computational resources during inference~\cite{brown2024large,chen2024more}, termed test-time scaling. These methods includes majority voting~\cite{wang2022self}, weighted aggregation~\cite{li2023making}, best-of-N~\cite{lightman2023let}, Tree-of-Thoughts~\cite{yao2023tree}, and Monte Carlo Tree Search variants~\cite{wu2024empirical,zhang2024accessing,zhao2024marco}. Besides, o1 model~\cite{jaech2024openai} and several follow-up works~\cite{guo2025deepseek,qwen2024qwq,google2025gemini,min2024imitate,huang2024o1} increase the thinking depth to improve the performance. But they all apply fixed scaling strategies to all questions. Some adaptive thinking methods like AdoT~\cite{xu2024adaption} and DAST~\cite{shen2025dast} design difficulty measurement to categorize the question based on its difficulty, whereas they introduce human bias and fail to leverage the model's inherent reasoning abilities. Our T$^2$ framework builds upon this paradigm while addressing these key limitations.

\section{T$^2$: Think-to-Think Framework}
\label{sec:methodology}

In this section, we present T$^2$: Think-to-Think, an approach that enables language models to adapt their reasoning strategies based on question complexity. Figure~\ref{fig:overview} provides an overview of our approach. We begin by describing the overall architecture and workflow of T$^2$ before delving into each component in detail. 


\subsection{Question Decomposition}
\label{sec:question-structure-identification}

Given a document $D$ and a question $Q$, we first analyze the question's structure to understand its underlying pattern. This allows us to later generate similar questions that require same reasoning strategy. The question structure identification process involves decomposing the question into fixed structural elements and variable entities that could be substituted.

We first tokenize the question $Q$ as a sequence of tokens $Q = (q_1, q_2, \ldots, q_m)$. We then classify each token into one of two categories: structural tokens that form the question's framework, and replaceable entities that could be substituted with alternatives. We define a classification function with fine-tuned RoBERTa, detailed in Appendix~\ref{apd:implementations}. Based on this classification, we partition the question tokens into two sets:
\begin{align}
    P &= \{q_i \mid \text{if $q_i$ is a replaceable entity}\}, \\
    Q_S &= \{q_i \mid \text{if $q_i$ is a structural token}\},
\end{align}
where $P$ represents the set of replaceable entities (which we call entity placeholders), and $Q_S$ represents the set of structural tokens that form the question's framework.


For each identified entity placeholder $p_i$ in $P$, we assign a semantic type (e.g., \texttt{person}, \texttt{location}, \texttt{date}). This creates a set of typed entities:
\begin{equation}
T = \{(p_1, \tau_1), (p_2, \tau_2), \ldots, (p_k, \tau_k)\},
\end{equation}
where each pair $(p_j, \tau_j)$ consists of a placeholder entity $p_j$ and its corresponding type $\tau_j$.

By combining the structure tokens $Q_S$ with the typed placeholders in $T$, we create a question template. For example, if $Q$ is ``Which is taller, the Eiffel Tower or the Empire State Building?'', the function would identify ``taller'', ``Eiffel Tower'', and ``Empire State Building'' as replaceable entities of type \texttt{adj} and \texttt{place}. The resulting template would be ``Which is [adj], [place 1] or [place 2]?'', where the bracketed terms are typed placeholders.

\subsection{Similar Examples Generation}
\label{sec:synthetic-qa-generation}

Once we have extracted the question structure, we generate similar document-question-answer pairs that follow the same question structure but with different entities.

\paragraph{Reasoning Skills Taxonomy.}
We build on established cognitive science literature \citep{sep-reasoning-analogy,774846} to define a taxonomy of 7 fundamental reasoning skills $\mathcal{S}$ that humans commonly employ when solving problems (e.g., \textit{Deductive}, \textit{Inductive}\footnote{Appendix~\ref{apd:reasoning_skills} shows the complete taxonomy of reasoning skills with their description and example applications.}). Each skill represents a distinct cognitive approach to processing information and drawing conclusions.

\paragraph{Question Generation.}
For each placeholder in the question structure, we generate alternative entities of matching types. We prompt an LLM to suggest contextually appropriate substitutes for each entity type $\tau_j$. This produces a collection of candidate similar questions $\hat{Q}_{\text{sim}}$ that share the structural pattern of the original question but contain different entities.

To ensure high-quality examples, we implement a validation process. We prompt the same LLM to evaluate the similarity between each candidate question and the original question structure:
\begin{equation}
\text{sim}(Q, \hat{q}) \geq \delta, \quad \hat{q} \in \hat{Q}_{\text{sim}},
\end{equation}
where $\delta \in [1, 10]$ is a threshold parameter. Only questions exceeding this threshold are retained, resulting in a filtered set of similar questions $Q_{\text{sim}}$.

\paragraph{Reasoning Strategy Construction.}
For each similar question $Q_{\text{sim}}^i\in Q_{\text{sim}}$, we decompose it into a sequence of subquestions:
\begin{equation}
Q_{\text{sim}}^i \rightarrow (Q_{\text{sim}}^{(i,1)}, \ldots, Q_{\text{sim}}^{(i,K)}),
\end{equation}
where each subquestion $Q_{\text{sim}}^{(i,K)}$ represents a discrete reasoning step and $K$ is the number of subquestions. The connections between subquestions are characterized by specific reasoning skills from our taxonomy. This decomposition allows us to construct a comprehensive reasoning strategy:
\begin{equation}
\mathbf{s}^i = (s_1^i, s_2^i, \ldots, s_{K}^i),
\end{equation}
where each $s_k^i \in \mathcal{S}$ is the reasoning skill required to transition from subquestion $Q_{\text{sim}}^{(i,k)}$ to $Q_{\text{sim}}^{(i,k+1)}$.

\paragraph{Reference Document Generation.}
For each subquestion $Q_{\text{sim}}^{(i,k)}$, we generate a document segment $d_k^i$ containing the precise information needed to answer that subquestion. The complete reference document for question $Q_{\text{sim}}^i$ is then constructed as:
\begin{equation}
D_{\text{ref}}^i = \{d_1^i, d_2^i, \ldots, d_K^i\}.
\end{equation}

For example, given a similar question like ``Which is taller, A or B?'', the decomposition might yield subquestions: ``What is the height of A?'', ``What is the height of B?'', and ``Which height is greater?''. The reasoning strategy would connect these using deductive reasoning, and the reference document would provide the necessary height information for both entities.

The complete collection of similar examples is represented as:
\begin{equation}
\Gamma = \{(D_{\text{ref}}^i, Q_{\text{sim}}^i, \mathbf{s}^i)\}_{i=1}^{N},
\end{equation}
where $N$ is the total number of similar examples. This diverse set covers various reasoning strategies of different complexity levels, allowing our system to later select the most appropriate reasoning approach for original questions.

\subsection{Multi-Criteria Matching}
\label{sec:demo-selection}

When presented with the original question $Q$ and documents $D$, we need to determine which reasoning strategy would be most effective. We select the most relevant example from our similar collection $\Gamma$ using a multi-criteria matching process that considers both reasoning skill requirements and structural similarity.

\paragraph{Skill Uniqueness Scoring.}
Recognizing that some reasoning skills are more specialized than others, we weight skills by their rarity in our example collection. For each reasoning skill $s \in \mathcal{S}$, we define $\text{freq}(s)$ as the number of examples in $\Gamma$ that include skill $s$ in their reasoning paths. The uniqueness score of a skill is:
\begin{equation}
\alpha(s) = \ln\left(\frac{N + 1}{\text{freq}(s) + 1}\right),
\end{equation}
where $N$ is the total number of examples in our collection. This logarithmic formulation assigns higher weights to skills that appear less frequently, capturing the intuition that specialized reasoning skills deserve special consideration.

\paragraph{Skill Coverage Assessment.}
For each example in our collection, we calculate how well its reasoning path covers reasoning skills:
\begin{equation}
\text{cover}(\mathbf{s}^i, \mathcal{S}) = \frac{|\mathbf{s}^i \cap \mathcal{S}|}{|\mathcal{S}|}.
\end{equation}

This coverage metric quantifies what proportion of the required reasoning skills are present in the example's reasoning strategy.

\begin{algorithm*}[!t]
\caption{Reasoning Path-Guided Answering}
\label{alg:answer-gen}
\begin{algorithmic}[1]
\Require $Q$ (original question), $D$ (document), $i^*$ (selected example index), $\Gamma$ (example collection)
\Ensure $A$ (final answer)

\State $(D_{\text{ref}}^{i^*}, Q_{\text{sim}}^{i^*}, A_{\text{sim}}^{i^*}, \mathbf{s}^{i^*}) \gets \Gamma[i^*]$ \Comment{Retrieve selected example}

\State $D_{\text{focus}} \gets \emptyset$ \Comment{Initialize focused document segments}

\For{$\ell = 1$ to $|\mathbf{s}^{i^*}|$} \Comment{For each skill in the reasoning path}
    \State $\text{text}_\ell \gets \text{ExtractRelevantSegment}(D, s_{\ell}^{i^*})$ \Comment{Extract relevant text for skill $s_{\ell}^{i^*}$}
    \State $D_{\text{focus}} \gets D_{\text{focus}} \cup \{\text{text}_\ell\}$ \Comment{Add to focused segments}
\EndFor

\State \text{Prompt} $\gets$ \text{FormatPrompt}$(Q, D_{\text{focus}}, \mathbf{s}^{i^*}, Q_{\text{sim}}^{i^*}, A_{\text{sim}}^{i^*})$ \Comment{Construct guidance prompt}

\State $A \gets \text{LLM}(\text{Prompt})$ \Comment{Generate answer with guided reasoning}

\State \textbf{return} $A$
\end{algorithmic}
\end{algorithm*}

\paragraph{Integrated Selection Score.}
We compute a comprehensive selection score for each remaining example, and the optimal example is selected as:
\begin{equation}
i^* = \argmax_i \left(\text{cover}(\mathbf{s}^i, \mathcal{S}) + \sum_{\ell=1}^L \alpha(s_\ell^i)\right),
\end{equation}
where $L$ is the length of the reasoning strategy $\mathbf{s}^i$. This score balances how well the example covers the required reasoning skills and how uniquely it captures specialized reasoning approaches.


\subsection{Reasoning Strategy-Guided Answering}
\label{sec:reasoning-path-guided-answering}

The final component of T$^2$ uses the selected example to guide the reasoning process for answering the original question. Algorithm~\ref{alg:answer-gen} outlines this process.

The ``ExtractRelevantSegment'' function uses LLM to identify portions of the document $D$ that are most relevant to applying a particular reasoning skill. This focuses the model's attention on information appropriate to each step of the reasoning process. The ``FormatPrompt'' function combines the original question, the focused document segments, the selected reasoning strategy, and the example document-question-answer pair into a comprehensive prompt. This prompt instructs the language model to answer the original question by applying the reasoning skills in the selected strategy, using the example as a demonstration of the reasoning approach.

This methodology enables adaptive reasoning that scales with question complexity. For simple questions, T$^2$ selects examples with a straightforward reasoning strategy, avoiding unnecessary computational overhead. For complex questions, it selects examples with a more sophisticated reasoning strategy that guides the model through the necessary steps to arrive at the correct answer. Importantly, this adaptation occurs without parameter tuning or multiple reasoning attempts, requiring only a single forward pass through the language model.

\section{Experiments}

\subsection{Experimental Setups}
\label{sec:experimental-setup}

\paragraph{Datasets.} We evaluate our approach on seven QA datasets from diverse domains. \textbf{SQuAD} (general-domain questions from Wikipedia)~\cite{rajpurkar-etal-2018-know}, \textbf{HotpotQA} (multihop questions spanning multiple paragraphs)~\cite{yang2018hotpotqa}, \textbf{BioASQ} (biomedical queries requiring specialized knowledge)~\cite{283}, \textbf{NewsQA} (news-related passages)~\cite{trischler2017newsqa}, \textbf{GAOKAO} (exam-oriented dataset with academic coverage)~\cite{zhang2024evaluatingperformancelargelanguage}, \textbf{HQA} (historical questions focusing on chronology and figures)~\cite{HOSEN2023109245}, and \textbf{TriviaQA} (Wikipedia-based trivia)~\cite{joshi-etal-2017-triviaqa}. Appendix~\ref{apd:datasets} summarizes dataset sizes and domains.

\begin{table*}[!t]
\centering
\small
\adjustbox{max width=\textwidth}{
\begin{tabular}{lccccccc}
\toprule
Model & SQuAD & HotpotQA & NewsQA & Gaokao & HQA & TriviaQA & BioASQ\\
\midrule
\rowcolor{gray!10}\multicolumn{8}{c}{\textbf{\textit{Quick-Thinking Models w/ Reasoning Strategies}}}\\
\multicolumn{8}{l}{\textbf{\textit{Qwen2.5-32B-Instruct}}}\\
\; w/ vanilla \textit{(quick)} & 73.41 & 55.32 & 50.83 & 29.52 & 35.92 & 40.73 & 56.33\\
\; w/ few-shots \textit{(quick)}  & 74.56 & 56.23 & 51.67 & 30.33 & 36.87 & 41.57 & 57.17\\
\; w/ self-consistency~\cite{wang2022self} & 75.31 & 56.76 & 52.27 & 30.57 & 37.12 & 41.92 & 57.57\\
\; w/ proCoT~\cite{deng-etal-2023-prompting} & 77.12 & 58.07 & 53.57 & 31.42 & 38.03 & 42.83 & 58.62\\
\; w/ ToT~\cite{yao2023tree}  & 78.47 & 59.11 & 54.31 & 31.96 & 38.66 & 43.46 & 59.36\\
\; w/ MCTS~\cite{zhao2024marco} & 78.52 & 58.97 & 54.25 & 32.04 & 38.73 & 43.51 & 59.42\\
\textbf{\; w/ T$^2$ (ours)} & \textbf{81.86} & \textbf{67.11} & \textbf{61.27} & \textbf{34.06} & \textbf{40.31} & \textbf{43.92} & \textbf{65.02}\\
\hdashline
\multicolumn{8}{l}{\textbf{\textit{GPT-4o}}}\\
\; w/ vanilla \textit{(quick)} & 78.52 & 60.02 & 55.32 & 34.51 & 41.11 & 49.01 & 60.51\\
\; w/ few-shots \textit{(quick)} & 79.86 & 61.06 & 56.17 & 35.36 & 42.06 & 50.07 & 61.37\\
\; w/ self-consistency~\cite{wang2022self} & 80.56 & 61.61 & 56.62 & 35.62 & 42.46 & 50.42 & 61.81\\
\; w/ proCoT~\cite{deng-etal-2023-prompting} & 82.12 & 63.02 & 57.86 & 36.66 & 43.36 & 51.46 & 62.87\\
\; w/ ToT~\cite{yao2023tree} & 83.21 & 64.06 & 58.67 & 37.22 & 44.07 & 52.26 & 63.72\\
\; w/ MCTS~\cite{zhao2024marco} & 83.35 & 64.18 & 58.19 & 37.31 & 45.15 & 52.38 & 64.89\\
\textbf{\; w/ T$^2$ (ours)} & \textbf{85.06} & \textbf{66.16} & \textbf{60.92} & \textbf{37.57} & \textbf{45.27} & \textbf{53.92} & \textbf{66.97}\\
\midrule
\rowcolor{gray!10}\multicolumn{8}{c}{\textbf{\textit{Slow-Thinking Models}}}\\
o1-mini                         & 85.81 & 70.91 & 63.22 & 42.66 & 49.22 & 58.56 & 68.42\\
QwQ-32B-Preview                     & 86.87 & 71.86 & 63.92 & 43.23 & 49.62 & 59.16 & 69.02\\
DeepSeek-R1                         & 87.62 & 72.72 & 64.41 & 43.47 & 50.27 & 60.02 & 70.72\\
o1                         & 88.22 & 73.37 & 65.11 & 44.06 & 51.07 & 60.86 & 71.36\\
o4-mini                              & 88.72 & 73.86 & 65.57 & 44.32 & 51.61 & 61.11 & 71.82\\
o4-mini-high                         & 88.91 & 74.07 & 65.81 & 44.52 & 51.86 & 61.27 & 72.02\\
Claude-3.7-sonnet-thinking          & 89.11 & 74.21 & 66.01 & 44.61 & 52.01 & 61.47 & 72.22\\
o3                              & 89.41 & 74.61 & 66.32 & 45.01 & 52.11 & 61.81 & 72.62\\
Gemini-2.5-Pro                      & 90.27 & 75.46 & 67.11 & 45.76 & 53.07 & 62.68 & 73.57\\
\textbf{QwQ-32B + T$^2$ (ours)}      & \textbf{92.12} & \textbf{77.61} & \textbf{68.61} & \textbf{47.42} & \textbf{54.71} & \textbf{64.22} & \textbf{75.21}\\
\bottomrule
\end{tabular}}
\caption{ROUGE-L on seven QA datasets. We regard vanilla model and few-shot method as quick-thinking methods. And the other five (including ours) are slow-thinking methods. They can all be applied to quick-thinking models to improve reasoning ability.}
\label{tab:main_results}
\end{table*}

\paragraph{Reasoning Strategies and Metrics.} We compare our T$^2$ framework against \textit{slow-thinking} and \textit{quick-thinking} baselines. Slow-thinking approaches include: \textbf{proactiveCoT (proCoT)}~\cite{deng-etal-2023-prompting}, \textbf{Self-Consistency}~\cite{wang2022self}, \textbf{Tree of Thoughts (ToT)}~\cite{yao2023tree}, and \textbf{Monte Carlo Tree Search (MCTS)}~\cite{zhao2024marco}. Quick-thinking methods include: \textbf{few-shot prompting} and \textbf{direct prompting} without explicit reasoning steps. For evaluation, we use ROUGE-L as our metric across all datasets.\footnote{We recognize GenAI can generate the correct answer, but with different literalness. Hence we use ROUGE-L here instead of resulting in a misleadingly low Exact Match (EM) rate. We also report EM performance in Appendix~\ref{apd:main_performance_em}.}

\paragraph{Large Language Models.} We use two quick-thinking LLMs (\textbf{Qwen2.5-32B-Instract}~\cite{qwen2.5}, and \textbf{GPT-4o}~\cite{hurst2024gpt,guo2025deepseek}) and several slow-thinking LLMs (\textbf{GPT-o1/3/4 series}~\cite{jaech2024openai}, \textbf{QwQ-32B-Preview}~\cite{qwq}, \textbf{Claude-3.7}~\cite{claude3-7}, \textbf{Gemini-2.5-Pro}~\cite{gemini2-5}). Unless otherwise specified, hyperparameters are set to the default values for each model. No domain-specific fine-tuning and no target-designed prompt are applied, ensuring a fair and consistent comparison. More detailed implementation and all prompts can be found in Appendices \ref{apd:implementations} and \ref{apd:prompts}.

\subsection{Results}
\label{sec:overall-perf}

Table~\ref{tab:main_results} compares ROUGE-L on seven QA benchmarks.  
The upper half lists \textit{quick-thinking models} evaluated with several slow-thinking frameworks. The lower half gathers the strongest \textit{slow-thinking models}. We also report the performance of Qwen2.5-32B-Instruct + T$^2$ and QwQ-32B-Preview + T$^2$ to show comparison with slow-thinking models. The experimental results show that by comparison with other thinking strategies, our T$^2$ could help quick-thinking model achieve better performance. Besides, compared with other slow-thinking models, adding our T$^2$ can also help model improve the performance. We conduct several analysis experiments detailed as follows.

\begin{figure*}[!t]
    \centering
    \includegraphics[width=\linewidth]{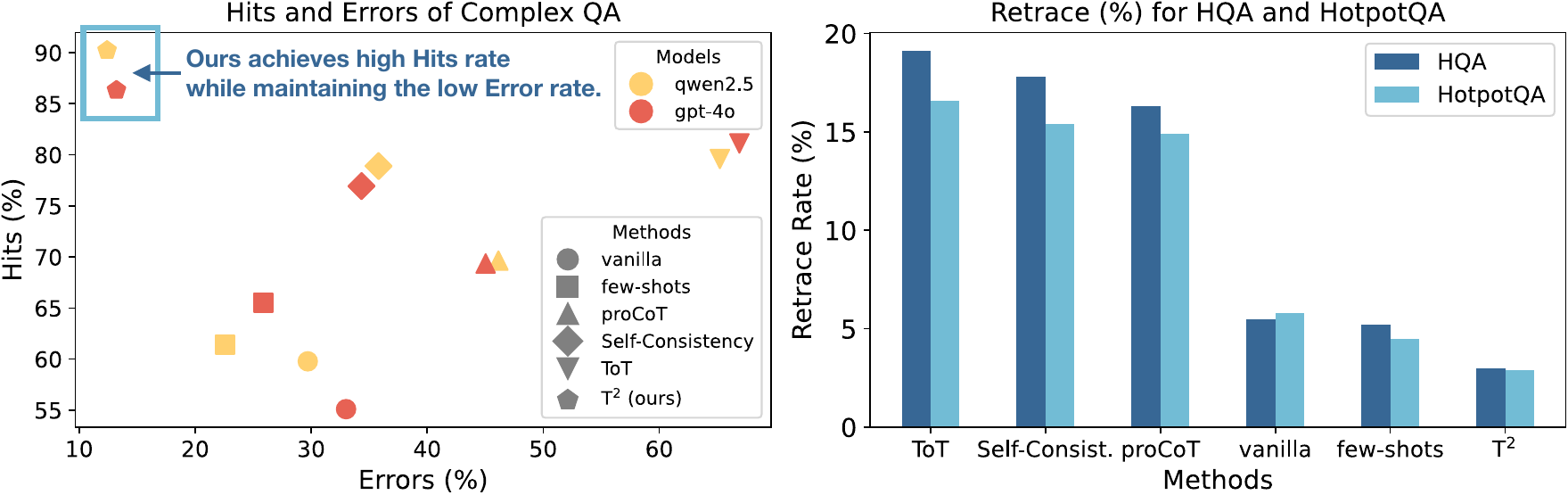}
    \caption{Results on Hits and Errors (left) and Retrace Rate (right).}
    \label{fig:hit_error}
    \label{fig:retrace}
\end{figure*}

\subsubsection{T$^2$ Enhance the Reasoning Skills Hit Rate while Reducing the Error}
\label{sec:effectiveness-multihop}

HotpotQA supplies gold supporting sentences for every question, hence we use these to evaluate reasoning quality. For a model output that mentions a set $P_q$ of sentences and a gold set $G_q$, we record a \emph{Hit} if $P_q\supseteq G_q$ (all required facts retrieved) and an \emph{Error} if $P_q\not\subseteq G_q$ (at least one spurious fact added). Thus Hit measures \emph{completeness}, Error measures \emph{precision}, and the two are inversely related: longer chains tend to raise Hit but also raise Error. Figure \ref{fig:hit_error}(left) shows that quick-thinking frameworks give low Hit and moderate Error, while slow-thinking methods improve Hit at the cost of higher Error. Our \textbf{T$^2$} strikes the best balance, achieving the highest Hit and the lowest Error on Qwen2.5-32B, confirming that adaptive path length yields the most accurate multihop reasoning. The detailed calculation of Hits and Errors can be found in Appendix~\ref{apd:hits_errors}.

\begin{table}[!t]
\centering
\adjustbox{max width=\linewidth}{
\begin{tabular}{cccc}
\toprule
\textbf{Skill Type} & \textbf{Uniform} & \textbf{Ours} & \textbf{Improvement} \\
\midrule
Deductive & 72.3\% & 75.8\% & +3.5\% \\
Inductive & 68.7\% & 73.2\% & +4.5\% \\
Abductive & 74.1\% & 76.3\% & +2.2\% \\
Cause \& Effect & 70.5\% & 74.1\% & +3.6\% \\
Analogical & 63.8\% & 71.5\% & +7.7\% \\
Critical Thinking & 69.2\% & 72.8\% & +3.6\% \\
Decompositional & 61.4\% & 69.7\% & +8.3\% \\
\bottomrule
\end{tabular}}
\caption{Performance comparison between uniform and our matching strategies.}
\label{tab:skill_match}
\end{table}

\subsubsection{T$^2$ Tends to Get Correct Answers Immediately without Retrace}
\label{sec:retrace-analysis}

A response is said to \emph{retrace} if the model announces a provisional conclusion and later back-tracks on it inside the same output (e.g., ``\emph{So the answer is X… wait, that seems wrong—let me revise… the answer is Y}''). Obviously, as retrace brings extra computing cost, it would be better for a model to ensure a lower retrace rate while maintaining the same accuracy. Concretely, we scan the CoT for either (i) \textit{<answer>} markers that appear more than once, or (ii) lexical repair cues such as ``sorry,'' ``actually,'' or ``let me rethink,'' followed by a different answer span; if either pattern occurs, the example counts as a retrace. Figure \ref{fig:retrace} (right) shows that, taking Qwen2.5-32B as LLM, slow-thinking methods retrace more on NewsQA and HQA, whereas quick-thinking methods seldom retrace but miss clues, hurting performance. Our \textbf{T$^2$} keeps both metrics low—matching the speed of quick thinking and the accuracy of slow thinking—demonstrating that adaptive path length minimises wasted reasoning. The detailed calculation of Hits and Errors can be found in Appendix~\ref{apd:retrace}.

\subsubsection{T$^2$ Costs Fewer Tokens to Achieve Superior Performance}

\begin{figure}[!t]
    \centering
    \includegraphics[width=\linewidth]{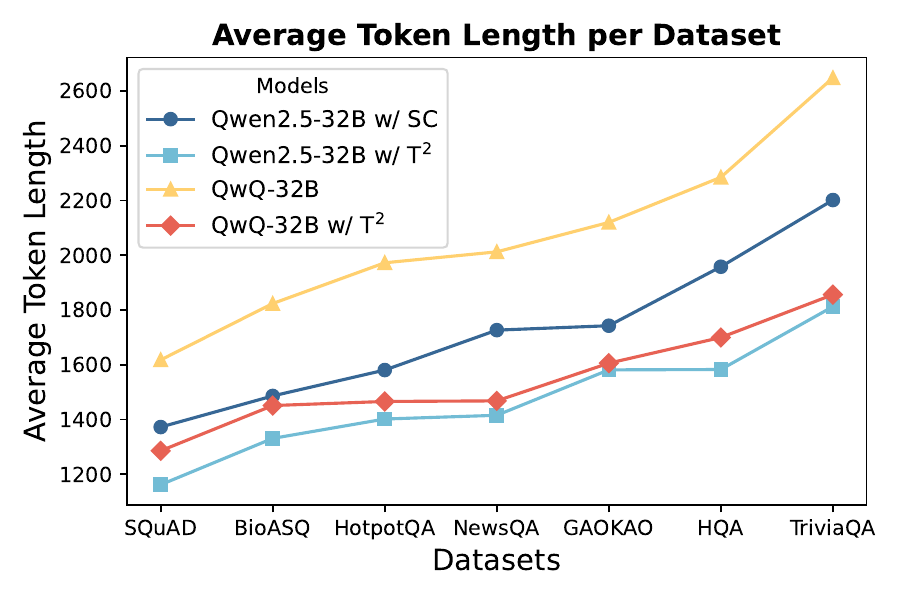}
    \caption{Results of average token length on each dataset. SC is the abbreviation for Self-Consistency.}
    \label{fig:ave_length}
\end{figure}

To evaluate the efficiency of our T$^2$, we compare four reasoning approaches: (1) {Qwen2.5-32B w/ self-consistency}, a typical slow-thinking method, (2) {QwQ-32B-Preview}, another slow-thinking model, (3) {Qwen2.5-32B w/ T$^2$}, and (4) {QwQ-32B w/ T$^2$}, our adaptive reasoning methods. Figure~\ref{fig:ave_length} shows that our method reduces token consumption by 25.2\% compared to QwQ-32B-Preview, and by 14.8\% compared to Qwen2.5-32B w/ self-consistency, while maintaining competitive accuracy. These findings highlight that our method achieves an optimal trade-off between computational efficiency and reasoning quality. A full comparison, including token usage and performance across datasets, is provided in Appendix~\ref{apd:efficiency_analysis}.

\begin{figure*}[!t]
    \centering
    \includegraphics[width=\linewidth]{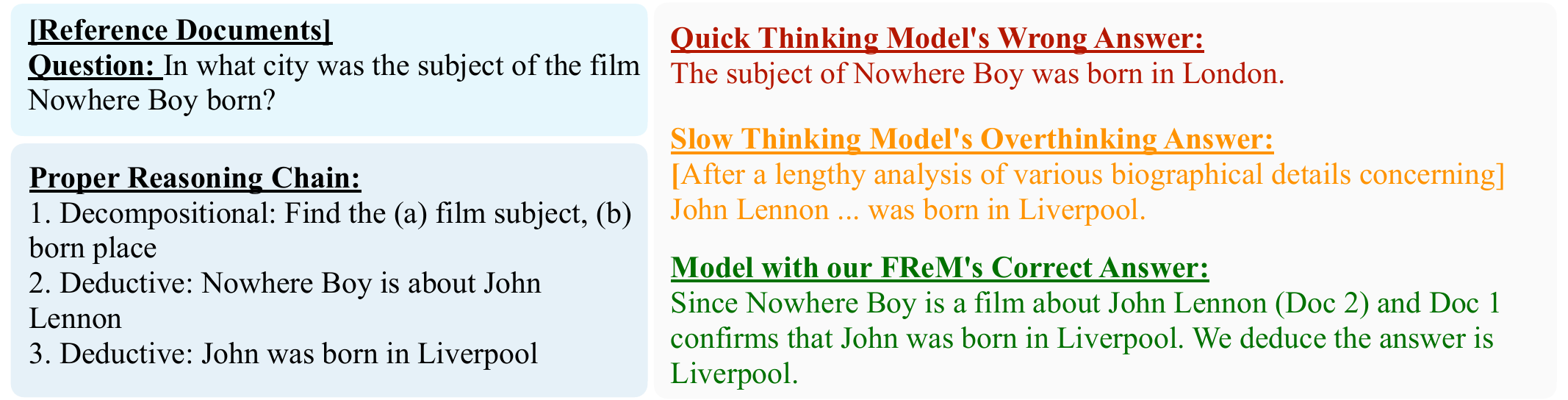}
    \caption{Case study to show effectiveness of our T$^2$ framework. There are three proper reasoning skills should be adopted to answer the question based on given documents. The red, orange, and green answers represent responses under quick thinking, slow thinking, and ours, respectively.}
    \label{fig:case}
\end{figure*}

\begin{figure}[!t]
    \centering
    \includegraphics[width=\linewidth]{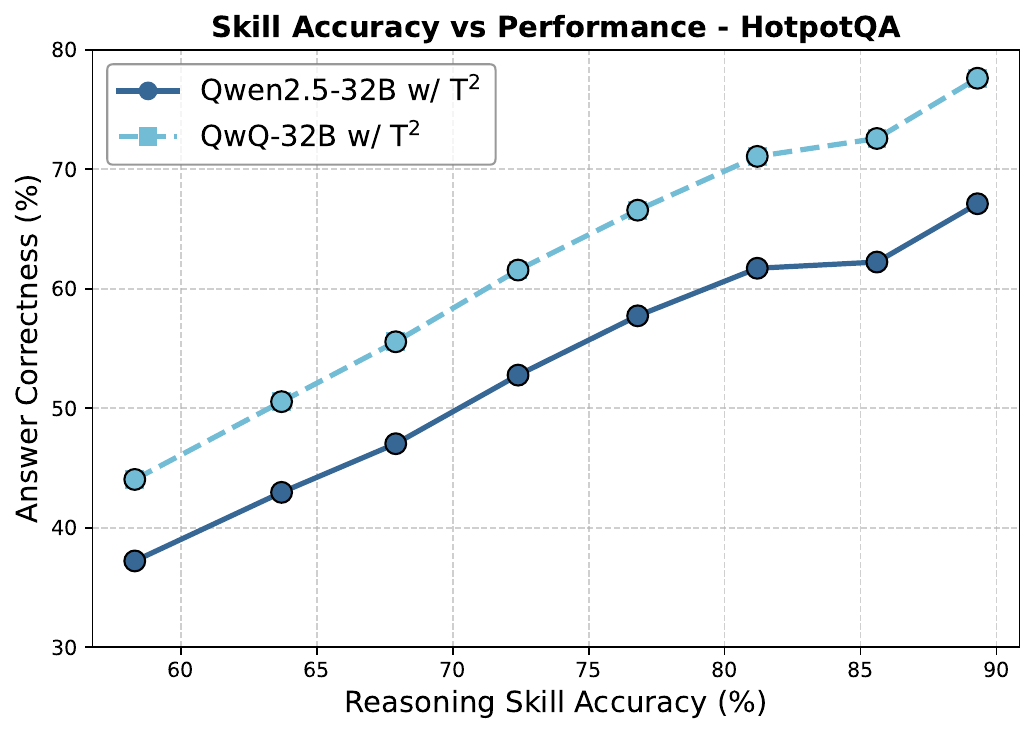}
    \caption{Results on relationship between reasoning skills' accuracy and overall performance.}
    \label{fig:demo_quality}
\end{figure}

\subsection{Similar Examples Quality Analysis}

\paragraph{Our Matching Strategy Can Expose More Diverse Reasoning Skills.} The effectiveness of our framework relies not only on identifying appropriate reasoning skills but also on how these skills are matched during the example selection. Hence, we examine the impact of our multi-criteria reasoning skills matching strategy compared to a naive uniform sampling approach. Table~\ref{tab:skill_match} presents the results of our experiment against uniform sampling across different reasoning skill types. Our approach consistently outperforms uniform sampling across all skill categories, with particularly notable improvements for less frequent reasoning types such as decompositional reasoning (+8.3\%) and analogical reasoning (+7.7\%). This confirms our hypothesis that the strategic balancing of skill demonstrations enhances the model's ability to leverage diverse reasoning patterns. The distribution of each reasoning skill can be found in Appendix~\ref{apd:skill_distribution}. Ablation study of multi-criteria matching strategy can be found in Appendix~\ref{apd:ablation_matching_strategy}.

\paragraph{Accuracy of Reasoning Skills Results in Correctness of Answers.} We examined the correlation between the accuracy of selected reasoning skills and the correctness of final answers using the HotpotQA dataset. We conducted the experiment on two models: Qwen2.5-32B-Instruct w/ T$^2$ and QwQ-32B-Preview w/ T$^2$. The analysis, shown in Figure \ref{fig:demo_quality}, reveals a strong positive correlation between skill accuracy and answer correctness. Higher skill accuracy corresponds to higher answer correctness, with an approximate 5-6\% increase in correctness for every 5\% improvement in skill accuracy. These results demonstrate that accurately selecting the correct reasoning skills is essential for generating correct answers, especially in complex multi-hop reasoning tasks.



We also discuss the impacts of question structure (\ref{apd:ipt_of_q_structure}), impacts of numbers of similar examples (\ref{apd:ipt_of_d_number}), impacts of various generated methods (\ref{apd:ipt_of_d_generate}), impacts of threshold of similarity in generation (\ref{apd:ipt_of_d_threshold}), impacts of examples domain bias and structural bias (\ref{apd:ipt_of_d_domain}), and human evaluation (\ref{apd:ipt_of_d_quality}) in Appendix. 

\subsection{Case Study}

Figure \ref{fig:case} shows an short version of example to show effectiveness of our T$^2$. By explicitly providing the model-specific reasoning path, the model can generate the correct answer with an appropriate reasoning chain of thought. The detailed case studies can be found in Appendix \ref{apd:case_studies}.

\section{Conclusion}

In this paper, we introduced T$^2$: Think-to-Think, a novel framework that dynamically adapts reasoning depth based on question complexity for contextual question answering tasks. Unlike prior approaches that employ fixed reasoning strategies regardless of question difficulty, T$^2$ enables models to learn appropriate reasoning strategies from similar examples, leading to more efficient processing while maintaining accuracy. Our experimental results across seven diverse CQA benchmarks confirm that T$^2$ not only achieves higher accuracy than baseline methods but also reduces computational overhead by up to 25.2\%. These improvements demonstrate the value of adaptability in reasoning processes, suggesting that as language models continue to evolve, approaches like T$^2$ that optimize both accuracy and computational efficiency will become increasingly important for developing more intelligent systems that can effectively allocate computational resources based on task demands.

\section*{Limitations}

While T$^2$: Think-to-Think demonstrates promising results across various CQA benchmarks, we acknowledge several limitations of our approach: First, the effectiveness of T$^2$ relies on the availability of high-quality example reasoning strategy for similarity matching. In domains with limited annotated examples or highly novel questions, the framework may struggle to identify appropriate reasoning patterns, potentially defaulting to less optimal strategies. Besides, our current implementation focuses primarily on textual reasoning tasks. Extending T$^2$ to multimodal reasoning contexts (e.g., visual question answering) would require additional architectural modifications to handle diverse input modalities while maintaining computational efficiency. 
Despite these limitations, we believe T$^2$ represents a significant step toward more adaptive and efficient reasoning systems that can intelligently allocate computational resources based on question complexity.

\section*{Ethical Considerations}

We ensure that all experiments are conducted using publicly available, ethically sourced datasets, adhering to privacy and intellectual property guidelines. We acknowledge the potential for biases in data and are committed to evaluating and mitigating any such biases in T$^2$.

\bibliography{custom}

\appendix

\section{Full Reasoning Skills}
\label{apd:reasoning_skills}

Defined by~\cite{sep-reasoning-analogy,774846}, reasoning can best be defined as the basic action of thinking in a sensible and rational way about something. Reasoning is the ability to assess things rationally by applying logic based on new or existing information when making a decision or solving a problem. Based on their conclusion, Tables \ref{tab:full_reasoning_skills} and \ref{tab:full_reasoning_skills_2} show the reasoning skills for answering a certain question.

\begin{table*}[!t]
    \centering
    \small
    \begin{tabular}{l p{0.4\textwidth} p{0.35\textwidth}}
    \toprule
    \textbf{Type of Reasoning} & \textbf{Detailed Description} & \textbf{Example} \\
    \midrule
    \textbf{Deductive} 
    & Deductive reasoning occurs when generalized statements apply to specific cases. These generalized statements are established and already proven, making specific cases easy to deduce. For example, all humans are mortals. Bill is a human, so Bill must be mortal. In this example the generalized, but proven, statement, ``all humans are mortals'' is what drives the reasoning.
    & \underline{Document:} All shapes with three sides are triangles. A certain figure here has exactly three sides.
      \newline
      \underline{Question:} What is this figure called?
      \newline
      \underline{Answer:} It is a triangle. All shapes with three sides are triangles, and this figure has three sides. So it must be a triangle. \\
    \midrule
    \textbf{Inductive} 
    & Inductive reasoning is similar to deductive reasoning in that they both draw a conclusion based on a statement. However, in inductive reasoning, the statement is likely but has not been proven. For example, roses usually bloom in spring. In spring, one can count on there being roses. Again, the difference is that this is likely but not proven to be 100\%. 
    & \underline{Document:} Every spring for the past ten years, wild roses in Green Valley have bloomed in late March. This spring is about to begin in Green Valley.  
      \newline
      \underline{Question:} Will the wild roses bloom in late March this year?
      \newline
      \underline{Answer:} It is likely they will bloom in late March, because they usually do, but it is not guaranteed. \\
    \midrule
    \textbf{Abductive}
    & Abductive reasoning is the act of making a conclusion based on what you already know. For example, if you see a plate of food still hot, but half-eaten, you can make the conclusion that the person eating that food is probably returning soon.
    & \underline{Document:} You notice a half-eaten sandwich and a still-hot cup of coffee on a café table. The seat feels warm, and a jacket is draped over the chair.
        \newline
        \underline{Question:} Has the person who was sitting here left permanently, or are they coming back soon?
        \newline
        \underline{Answer:} It is likely they just stepped away for a moment and will return, because the food and drink are still warm and their jacket remains on the chair.\\
    \midrule
    \textbf{Cause \& Effect}
    & Cause and effect reasoning is that if x happens then y will happen as a result. This is extremely persuasive when making a speech or trying to get someone to take action to cause an effect. For example, a politician may say that if they are elected, then poverty will decrease. This is using cause and effect reasoning in a real-world situation.
    & \underline{Document:} Meteorologists predict heavy rain this evening, with warnings that streets may flood if the rainfall continues.
    \newline
    \underline{Question:} Will the roads become dangerous as a result of this weather?
    \newline
    \underline{Answer:} Yes. If heavy rain continues, roads will likely flood and become slippery, causing drivers to have less control of their vehicles.\\
    \bottomrule
    \end{tabular}
    \caption{(1/2) Full list of reasoning skills used in the reasoning path construction.}
    \label{tab:full_reasoning_skills}
\end{table*}

\begin{table*}[!t]
    \centering
    \small
    \begin{tabular}{l p{0.4\textwidth} p{0.35\textwidth}}
    \toprule
    \textbf{Type of Reasoning} & \textbf{Detailed Description} & \textbf{Example} \\
    \midrule
    \textbf{Analogical}
    & Analogical reasoning is the use of a comparison between two things to persuade that there must be more in common if they already share something. For example, if x, y, and z all share this trait, then they must also share other traits. The foundation of this type of reasoning is perfect for speeches and comparisons in the real world. If there are connections between x and y already, then they must have several other things in common as well.
    & \underline{Document:} Many leading technology companies emphasize continuous learning and adaptability. For instance, Google, Microsoft, and Amazon all invest in regular training programs and encourage innovation among employees. Their similar approach to fostering a culture of growth has been linked to their strong performance in rapidly changing markets.
    \newline
    \underline{Question:} Can we infer that a company that promotes continuous learning will also likely be successful in adapting to market changes?
    \newline
    \underline{Answer:} Yes. Since Google, Microsoft, and Amazon all share a culture of continuous learning and, as a result, demonstrate high adaptability and market success, it is reasonable to conclude by analogy that a company which also promotes continuous learning is likely to develop similar strengths.\\
    \midrule
    \textbf{Critical Thinking}
    & Critical thinking occurs when you take all of the facts and develop a conclusion based on an analysis. This could happen subconsciously or intentionally, depending on the situation. For example, in the real world, critical thinking could be about your relationships. You could see a behavior you don't like about someone and have to think critically about whether or not you will choose to spend more time with this person. This is using critical thinking to develop reasoning in a real-world application.
    & \underline{Document:} Over the past few months, Sam has repeatedly cancelled plans at the last minute and rarely communicated afterward.
    \newline
    \underline{Question:} Should you invest time in a close friendship with Sam?
    \newline
    \underline{Answer:} No. Sam's consistent behavior of last-minute cancellations suggests a pattern of unreliability, which may negatively affect the trust needed in a close friendship.\\
    \midrule
    \textbf{Decompositional}
    & Decompositional reasoning happens when the different parts of the reasoning are broken down into smaller pieces and analyzed for how they contribute to the whole. The intent of this is to make the reasoning easier to understand and allow for analyzing how the parts equal the whole. For example, in order to understand the function of the human body, you would have to analyze each bone and organ to see how they all work together. Additionally, in the real world, an argument could be broken down into several smaller parts in order to analyze the effectiveness of the argument as a whole.
    & \underline{Document:} A smartphone's quality can be understood by breaking it down into three parts: its design, performance, and battery life. The design covers the build and user interface; performance looks at processing speed and software efficiency; battery life shows how long the device operates on a single charge.
    \newline
    \underline{Question:} Can we conclude that the smartphone provides a good overall user experience?
    \newline
    \underline{Answer:} Yes. If the design is appealing, the performance is robust, and the battery life is long, then the smartphone is likely to offer a good overall experience.\\
    \bottomrule
    \end{tabular}
    \caption{(2/2) Full list of reasoning skills used in the reasoning path construction.}
    \label{tab:full_reasoning_skills_2}
\end{table*}

\section{Datasets}
\label{apd:datasets}

In this work, we evaluate our method on seven widely used question answering datasets. Each dataset presents distinct characteristics, ranging from the type of questions asked to the domain in which they are applied. Below, we provide a brief overview of each dataset.

\paragraph{SQuAD} consists of over 100,000 question-answer pairs derived from a set of Wikipedia articles. The task is to find the span of text that answers the question. SQuAD is widely used for evaluating machine reading comprehension models. The dataset includes two versions: SQuAD 1.1, which contains answerable questions, and SQuAD 2.0, which also includes unanswerable questions, making it more challenging. We use 2.0 version here.

\paragraph{HotpotQA} is a large-scale, multi-hop question answering dataset that requires reasoning across multiple supporting facts. The dataset includes over 113,000 question-answer pairs spanning various domains, where answers cannot be found in a single sentence or passage but require combining information from several documents. The questions in HotpotQA require a more complex reasoning process compared to typical single-hop datasets.

\paragraph{BioASQ} is a biomedical question answering dataset that provides information from scientific articles, primarily in the domain of biomedicine. It includes both factoid and complex questions that require understanding of scientific literature. BioASQ focuses on answering clinical, biomedical, and molecular biology-related questions using both structured and unstructured data sources.

\paragraph{NewsQA} is a dataset designed for reading comprehension tasks. It consists of over 100,000 question-answer pairs derived from news articles. The challenge of NewsQA lies in answering questions about real-world events from unstructured news stories, requiring models to handle various linguistic phenomena such as coreference, reasoning, and implicit understanding.

\paragraph{GAOKAO} is a dataset derived from the Chinese college entrance exam, also known as the "Gaokao". It contains questions related to various subjects, including Chinese literature, mathematics, and English. The questions in GAOKAO require both general knowledge and reasoning to answer. This dataset is specifically designed for the Chinese education system and is widely used in academic and educational research in China.

\paragraph{HQA} is a human-annotated dataset specifically designed for complex, open-domain question answering. It contains questions that require deep contextual understanding and can involve reasoning across long documents. The dataset includes various types of questions and answers across diverse domains, and it was created to test models' ability to perform reasoning tasks in realistic, open-ended settings.

\paragraph{TriviaQA} is a large-scale dataset that focuses on answering trivia questions, where each question is associated with a corresponding set of supporting documents. TriviaQA contains over 650,000 question-answer pairs sourced from trivia websites and requires models to retrieve relevant information from the documents and answer based on the provided facts. The dataset has questions spanning various topics such as history, geography, and general knowledge.

\section{Distribution of Reasoning Skills in Each Dataset}
\label{apd:skill_distribution}

Table~\ref{tab:skill_distribution} demonstrates the distribution of seven reasoning skills in different datasets. The variance in skill distribution highlights why our multi-criteria matching approach is crucial. Without it, high-frequency skills like deductive reasoning would dominate the demonstrations, while valuable but less common skills like abductive reasoning would be underrepresented.

\begin{table*}[!t]
\centering
\adjustbox{max width=\textwidth}{
\begin{tabular}{cccccccc}
\toprule
\textbf{Skill Type} & \textbf{SQuAD} & \textbf{HotpotQA} & \textbf{NewsQA} & \textbf{GAOKAO} & \textbf{HQA} & \textbf{TriviaQA} & \textbf{BioASQ} \\
\midrule
Deductive & 0.31 & 0.22 & 0.28 & 0.15 & 0.42 & 0.18 & 0.25 \\
Inductive & 0.23 & 0.18 & 0.15 & 0.12 & 0.13 & 0.21 & 0.19 \\
Abductive & 0.05 & 0.12 & 0.08 & 0.21 & 0.09 & 0.15 & 0.11 \\
Cause \& Effect & 0.12 & 0.15 & 0.13 & 0.22 & 0.08 & 0.19 & 0.14 \\
Analogical & 0.08 & 0.13 & 0.09 & 0.07 & 0.11 & 0.12 & 0.16 \\
Critical Thinking & 0.14 & 0.16 & 0.18 & 0.14 & 0.13 & 0.09 & 0.10 \\
Decompositional & 0.07 & 0.04 & 0.09 & 0.09 & 0.04 & 0.06 & 0.05 \\
\bottomrule
\end{tabular}}
\caption{Distribution (\%) of reasoning skills across benchmark datasets, showing the proportion of questions requiring each skill type.}
\label{tab:skill_distribution}
\end{table*}

\section{Implementations}
\label{apd:implementations}

We use a simple pretrained language model RoBERTa from Huggingface for detecting named entities or key numbers in the question to obtain the question structure. This classification task involves processing the input question to identify whether it contains a named entity or key number and assigning a type to the detected entity. The model performs this task by outputting binary labels (entity: Yes/No) first, and then the associated entity types (e.g., Person, Location, Date, Organization, Number, etc.).

\begin{table*}[!t]
    \centering
    \begin{tabular}{ll}
    \toprule
    \textbf{Parameter} & \textbf{Value} \\
    \midrule
    \textbf{Model} & RoBERTa\\
    \textbf{Full Name} & \texttt{FacebookAI/xlm-roberta-large-finetuned-conll03-english}\\
    \textbf{Batch Size} & 128 \\
    \textbf{Learning Rate} & 2e-5 \\
    \textbf{Optimizer} & AdamW \\
    \textbf{Dropout Rate} & 0.1 \\
    \textbf{Evaluation Metric} & Accuracy \\
    \bottomrule
    \end{tabular}
    \caption{Implementation parameters for named entity detection and classification.}
    \label{tab:implementation_parameters}
\end{table*}

This model is fine-tuned with a simple classification layer that detects whether a named entity or key number is present in the question with NERetrieve dataset\footnote{\url{https://github.com/katzurik/NERetrieve?tab=readme-ov-file}} \cite{katz2023neretrieve}. This process leverages the model's pre-trained knowledge, with minimal fine-tuning specifically focused on the entity detection and classification task.

The hyperparameters used for fine-tuning the PLM are listed in Table \ref{tab:implementation_parameters}. The batch size is set to 128. The learning rate is set to \(2 \times 10^{-5}\). AdamW is used as the optimizer. A dropout rate of 0.1 is applied to prevent overfitting during fine-tuning.

For LLM usage, We use two quick-thinking LLMs (\textbf{Qwen2.5-32B-Instract}~\cite{qwen2.5}, and \textbf{GPT-4o}~\cite{hurst2024gpt,guo2025deepseek}) and several slow-thinking LLMs (\textbf{GPT-o1/3/4 series}~\cite{jaech2024openai}, \textbf{QwQ-32B-Preview}~\cite{qwq}, \textbf{Claude-3.7}~\cite{claude3-7}, \textbf{Gemini-2.5-Pro}~\cite{gemini2-5}). For ToT implementation, we follow the original paper's approach~\cite{yao2023tree} with a breadth-first search strategy and a maximum depth of 3. For MCTS, we implement the standard UCT algorithm with 10 simulations per decision point. For synthetic QA generation, we set a maximum output length of 4,096 tokens. When deciding which similar example to use, we follow our multi-criteria matching (Section \ref{sec:demo-selection}) to pick the most relevant chain of skills. Unless otherwise specified, hyperparameters stay at default values for each model. No domain-specific fine-tuning and no targeted designed prompt are applied, ensuring a fair and consistent comparison. All inferences are based on vLLM framework.

\section{Inference Prompts}
\label{apd:prompts}

The primary task is to generate synthetic question-answer pairs with a reasoning path, reflecting predefined reasoning skills. Table \ref{tab:synthetic_QA_generation_prompt} shows our prompts.

\begin{table*}[!t]
    \centering
    \small
    \begin{tabular}{|l|}
    \hline
    \textbf{Prompt:} \\
    \text{You are a language model that generates synthetic question-answer (QA) pairs with reasoning paths.}\\
    \text{Your task is to generate a QA pair based on the following question. Additionally, you should}\\
    \text{provide a clear, step-by-step reasoning path that corresponds to a predefined reasoning skill.}\\
    \text{The predefined reasoning skills are:[REASONING SKILLS NAME+DESCRIPTION+EXAMPLES].}\\
    \text{Your reasoning path should include clear substeps for each step of the thought process.} \\
    \textbf{Example 1:}\\
    \textbf{Given Documents:} [REFERENCE DOCUMENTS]\\
    \textbf{Input Question:} "Who invented the telephone?" \\
    \textbf{Step-by-step Reasoning Path:} \\
    1. Identify the key entity: "telephone" (deductive)\\
    2. Identify that the question is asking for the inventor of a significant historical device (decompositional)\\
    3. Recall the historical context of the invention of the telephone. (deductive)\\
    4. The inventor is Alexander Graham Bell. (cause \& effect)\\
    \textbf{Generated Answer:} "Alexander Graham Bell invented the telephone in 1876." \\
    \textbf{Reasoning Skill Used:} deductive, decompositional, deductive, cause \& effect. \\
    \textbf{Example 2:} ...\\
    \textbf{Example 3:} ...\\
    \textbf{Notes:}\\
    \text{Please make sure that the reasoning path is clear and includes each substep in the thought process.}\\
    \text{The output should follow this structure: "Step-by-step reasoning," followed by the conclusion.}\\
    \text{Each reasoning skill corresponds to a specific domain of knowledge.} \\
    \hline
    \end{tabular}
    \caption{Prompt to Generate Similar Examples with Reasoning Paths.}
    \label{tab:synthetic_QA_generation_prompt}
\end{table*}

Table \ref{tab:alignment_prompt} shows the helper language model for evaluating how well each example's question aligns with original one.

\begin{table*}[!t]
\centering
\small
\begin{tabular}{|p{\textwidth}|}
\hline
\textbf{Prompt:}\\
{You are given an original question: [ORIGINAL QUESTION]}\\
{You also have a synthetic question: [SYNTHETIC QUESTION]}\\
{Your task is to decide how similar the synthetic question is in structure and complexity, compared to the original. Please provide a brief explanation of your reasoning. Then, assign a score from 1 (completely different) to 10 (very similar).}\\
\textbf{Example}:
{Original Q: "Who discovered penicillin?"}\\
{Synthetic Q: "Which scientist found the mold that led to antibiotics?"}\\
\textbf{Explanation}:
{Both questions ask about a discoverer of a major medical breakthrough. The second question focuses on the mold (penicillin), so it is structurally similar and retains the core inquiry about a discovery.}\\
\textbf{Score (1-10)}: {8}\\
\textbf{Notes}:
{- Provide a short justification.}\\
{- Avoid rewriting or changing the question.}\\
{- Keep the final output concise, ending with the numeric score.}\\
\hline
\end{tabular}
\caption{Prompt for Evaluating Alignment of Synthetic Questions with the Original.}
\label{tab:alignment_prompt}
\end{table*}

Table \ref{tab:reasoning_path_prompt} shows the question answering prompts for model.

\begin{table*}[!t]
\centering
\small
\begin{tabular}{|p{\textwidth}|}
\hline
\textbf{Prompt:}\\
{You are given:}\\
{- The original question: [Q]}\\
{- A document or context: [D]}\\
{- A selected reasoning path: [R]}\\
{- The specific skills used in the reasoning path: [S]}\\

\textbf{Your goal} is to produce a final answer by combining the relevant information from [D] with the guided reasoning steps from [R]. Follow these instructions:\\
1. \textbf{Review the Reasoning Path}\\
   Read each step in [R] carefully. Identify which parts of [D] or background knowledge support each step.  \\
2. \textbf{Apply the Skills}  \\
   If [S] includes certain reasoning skills (e.g., deduction), make sure to explicitly use them when combining evidence from [D].\\
3. \textbf{Generate a Clear Answer}  \\
   Compose a concise final answer that directly addresses [Q]. You may outline your chain of thought, but keep the explanation aligned with [R].  \\
4. \textbf{Maintain Accuracy}  \\
   If [R] instructs a specific substep (e.g., numerical calculation or bridging multiple facts), follow it precisely, citing the relevant parts of [D].\\
\textbf{Notes}:  \\
- {Do not contradict the provided reasoning path.}  \\
- {Cite relevant text from [D] if needed, but avoid unnecessary repetition.}  \\
- {End with a concise, standalone final answer.}  \\
\hline
\end{tabular}
\caption{Prompt for Question Answering.}
\label{tab:reasoning_path_prompt}
\end{table*}

\section{Performance with Exact Match Metric}
\label{apd:main_performance_em}

\begin{table*}[!t]
\centering
\small
\adjustbox{max width=\textwidth}{
\begin{tabular}{lccccccc}
\toprule
Model & SQuAD & HotpotQA & NewsQA & Gaokao & HQA & TriviaQA & BioASQ\\
\midrule
\rowcolor{gray!10}\multicolumn{8}{c}{\textbf{\textit{Quick-Thinking Models w/ Reasoning Strategies (Exact Match)}}}\\
\multicolumn{8}{l}{\textbf{\textit{Qwen2.5-32B-Instruct}}}\\
\; w/ vanilla \textit{(quick)} & 55.23 & 31.69 & 27.15 & 12.61 & 19.47 & 23.82 & 40.88\\
\; w/ few-shots \textit{(quick)}  & 56.42 & 32.35 & 28.13 & 13.06 & 20.19 & 24.58 & 41.76\\
\; w/ self-consistency~\cite{wang2022self} & 57.08 & 32.81 & 28.49 & 13.29 & 20.37 & 24.89 & 41.93\\
\; w/ proCoT~\cite{deng-etal-2023-prompting} & 58.65 & 33.81 & 29.32 & 13.86 & 21.02 & 25.58 & 42.77\\
\; w/ ToT~\cite{yao2023tree}  & 59.83 & 34.67 & 29.81 & 14.12 & 21.51 & 26.04 & 43.28\\
\; w/ MCTS~\cite{zhao2024marco} & 59.87 & 34.53 & 29.76 & 14.21 & 21.57 & 26.08 & 43.33\\
\textbf{\; w/ T$^2$ (ours)} & \textbf{62.65} & \textbf{39.98} & \textbf{34.12} & \textbf{15.72} & \textbf{22.58} & \textbf{26.43} & \textbf{48.14}\\
\hdashline
\multicolumn{8}{l}{\textbf{\textit{GPT-4o}}}\\
\; w/ vanilla \textit{(quick)} & 59.87 & 35.31 & 30.27 & 16.23 & 23.39 & 29.84 & 44.12\\
\; w/ few-shots \textit{(quick)} & 61.09 & 36.04 & 30.85 & 16.78 & 24.03 & 30.52 & 44.84\\
\; w/ self-consistency~\cite{wang2022self} & 61.63 & 36.42 & 31.14 & 16.92 & 24.28 & 30.78 & 45.13\\
\; w/ proCoT~\cite{deng-etal-2023-prompting} & 62.89 & 37.29 & 31.92 & 17.49 & 25.01 & 31.43 & 46.05\\
\; w/ ToT~\cite{yao2023tree} & 63.77 & 37.94 & 32.43 & 17.79 & 25.57 & 32.11 & 46.61\\
\; w/ MCTS~\cite{zhao2024marco} & 63.94 & 38.13 & 32.06 & 17.84 & 26.22 & 32.18 & 47.39\\
\textbf{\; w/ T$^2$ (ours)} & \textbf{65.17} & \textbf{39.51} & \textbf{33.76} & \textbf{17.98} & \textbf{26.31} & \textbf{33.12} & \textbf{49.07}\\
\midrule
\rowcolor{gray!10}\multicolumn{8}{c}{\textbf{\textit{Slow-Thinking Models (Exact Match)}}}\\
o1-mini                         & 65.82 & 42.89 & 35.08 & 21.87 & 29.51 & 36.52 & 50.68\\
QwQ-32B-Preview                     & 66.67 & 43.57 & 35.43 & 22.18 & 29.88 & 36.87 & 51.21\\
DeepSeek-R1                         & 67.38 & 44.04 & 35.94 & 22.28 & 30.26 & 37.52 & 52.31\\
o1                         & 67.92 & 44.57 & 36.42 & 22.53 & 30.75 & 38.12 & 52.94\\
o4-mini                              & 68.36 & 44.97 & 36.83 & 22.79 & 31.08 & 38.28 & 53.14\\
o4-mini-high                         & 68.54 & 45.18 & 37.01 & 22.89 & 31.23 & 38.42 & 53.27\\
Claude-3.7-sonnet-thinking          & 68.67 & 45.27 & 37.14 & 22.97 & 31.32 & 38.56 & 53.41\\
o3                              & 68.89 & 45.48 & 37.34 & 23.19 & 31.37 & 38.78 & 53.74\\
Gemini-2.5-Pro                      & 69.69 & 46.08 & 37.97 & 23.51 & 32.01 & 39.43 & 54.45\\
\textbf{QwQ-32B + T$^2$ (ours)}      & \textbf{71.32} & \textbf{47.87} & \textbf{39.17} & \textbf{24.63} & \textbf{33.28} & \textbf{40.39} & \textbf{55.81}\\
\bottomrule
\end{tabular}}
\caption{Exact Match (EM) scores on seven QA datasets.}
\label{tab:overall_performance_em}
\end{table*}

Generally, Open QA datasets use Exact Match as their metrics for evaluation. But in generative AI system, the models can generate correct answers but with different literalness (e.g., ``San Francisco'' and ``The San Francisco City'' and ``SF U.S.''). Hence we use ROUGE-L as metric in our overall performance evaluation. Besides, we also report our experimental results on EM in Table~\ref{tab:overall_performance_em}.

\section{Calculation of Proposed Metrics}

\subsection{Hits and Errors}
\label{apd:hits_errors}

\paragraph{Hits Metric Calculation.} To evaluate the quality of reasoning and fact retrieval in the generated outputs, we employ the Hits metric based on the gold supporting sentences provided in HotpotQA. For each question $q$, let $P_q$ represent the set of sentences mentioned in the model's reasoning process and $G_q$ denote the set of gold supporting sentences.

We calculate the Hits metric as follows:
\begin{equation}
\text{Hits} = \frac{\sum_{q \in Q} \mathbf{1}[P_q \supseteq G_q]}{|Q|}
\end{equation}

where $\mathbf{1}[\cdot]$ is an indicator function that equals 1 when the condition is satisfied and 0 otherwise, and $|Q|$ is the total number of questions in the evaluation set. This formulation is similar to recall in traditional information retrieval, measuring the proportion of questions for which all required facts were successfully retrieved.

\paragraph{Errorss Metric Calculation.} For the Error metric, we adopt the False Discovery Rate (FDR) formulation:
\begin{equation}
\text{Error} = \frac{\sum_{q \in Q} \mathbf{1}[P_q \not\subseteq G_q]}{\sum_{q \in Q} (\mathbf{1}[P_q \supseteq G_q] + \mathbf{1}[P_q \not\subseteq G_q])}
\end{equation}

This represents the proportion of spurious facts (false positives) among all retrieved facts, consistent with the FDR calculation as FP/(TP+FP).

These complementary metrics create a natural trade-off: longer reasoning chains tend to improve Hits by including more supporting facts but often at the expense of increasing Error through the introduction of irrelevant information. An ideal reasoning process would maximize Hits while minimizing Error, indicating that the model precisely identifies all necessary supporting facts without including extraneous information.

\subsection{Retrace Rate}
\label{apd:retrace}

We define a response as exhibiting a \emph{retrace} when the model initially states a provisional conclusion and subsequently revises it within the same output. This occurs in patterns such as ``\emph{So the answer is X... wait, that seems wrong—let me revise... the answer is Y}.'' To systematically identify retraces, we analyze the Chain-of-Thought (CoT) reasoning for two specific patterns: (i) multiple occurrences of \textit{<answer>} markers, or (ii) lexical repair cues (e.g., ``sorry,'' ``actually,'' ``let me rethink'') followed by a different answer span. If either pattern is detected, we count the example as containing a retrace.

The Retrace Rate is calculated as:
\begin{equation}
\text{Retrace Rate} = \\
\frac{\sum_{q \in Q} \mathbf{1}[\text{retrace detected in } q]}{|Q|}
\end{equation}

where $\mathbf{1}[\cdot]$ is an indicator function that equals 1 when a retrace is detected and 0 otherwise, and $|Q|$ is the total number of questions in the evaluation set. This metric quantifies the proportion of responses where the model explicitly revises its reasoning path, providing insight into the model's self-correction capabilities during the reasoning process.

\section{Ablation Study}
\label{apd:ablation_matching_strategy}

To validate the effectiveness of our multi-criteria matching approach, we conducted ablation studies by systematically removing or modifying key components of our selection mechanism.

\paragraph{Impact of Selection Components.}
We evaluated four variants of our selection approach: (1) using only skill coverage without uniqueness weighting, (2) using only skill uniqueness without coverage assessment, (3) using random selection from examples passing the similarity threshold, and (4) our full approach. The experiments are conducted on Qwen2.5-32B as LLM. Table~\ref{tab:ablation} shows performance across test sets.

\begin{table}[!t]
\centering
\adjustbox{max width=\linewidth}{
\begin{tabular}{lccc}
\toprule
\textbf{Method} & \textbf{HotpotQA} & \textbf{NewsQA} & \textbf{HQA} \\
\midrule
Random      & 32.4 & 38.7 & 19.2 \\
Coverage Only   & 41.6 & 46.3 & 27.8 \\
Uniqueness Only & 49.2 & 54.5 & 35.7 \\
Full Approach         & \textbf{67.1} & \textbf{61.3} & \textbf{40.3} \\
\bottomrule
\end{tabular}}
\caption{Ablation study results showing the impact of different components in our selection approach. We report ROUGE-L (\%) on three benchmark datasets.}
\label{tab:ablation}
\end{table}

Results demonstrate that while both skill coverage and uniqueness contribute positively to performance, their combination in our full approach produces the strongest results across all datasets, yielding improvements of 25.5\% over using only individual components.

\section{Efficiency Analysis}
\label{apd:efficiency_analysis}

This section provides a comprehensive analysis of the computational efficiency of our proposed Flexible Reasoning Method (T$^2$) in comparison to other reasoning approaches. We analyze both token consumption and performance across seven diverse question answering datasets.

\subsection{Token Consumption Analysis}

Table \ref{tab:token_consumption} presents the average token consumption of different reasoning approaches across seven CQA datasets. The token length directly correlates with the computational resources required and inference time. Our results indicate that T$^2$ consistently reduces token consumption while maintaining or improving performance compared to other reasoning methods.

\begin{table*}[!t]
\centering
\adjustbox{max width=\linewidth}{
\begin{tabular}{lccccccc}
\toprule
\textbf{Model} & \textbf{SQuAD} & \textbf{BioASQ} & \textbf{HotpotQA} & \textbf{NewsQA} & \textbf{GAOKAO} & \textbf{HQA} & \textbf{TriviaQA} \\
\midrule
Qwen2.5-32B w/ SC & 1372.18 & 1726.32 & 1485.87 & 2201.65 & 1957.93 & 1580.43 & 1742.41 \\
Qwen2.5-32B w/ T$^2$ & 1161.42 & 1401.52 & 1330.71 & 1812.28 & 1581.14 & 1415.18 & 1582.42 \\
\hdashline
QwQ-32B-Preview & 1617.42 & 2012.33 & 1823.49 & 2648.12 & 2284.80 & 1972.37 & 2119.88 \\
QwQ-32B w/ T$^2$ & 1285.36 & 1467.85 & 1450.75 & 1855.89 & 1699.45 & 1465.68 & 1605.56 \\
\bottomrule
\end{tabular}}
\caption{Average token consumption across seven CQA datasets for different reasoning approaches.}
\label{tab:token_consumption}
\end{table*}

\subsection{Efficiency-Performance Trade-off}

Table \ref{tab:efficiency_performance} presents a comprehensive comparison of computational efficiency and performance across all seven datasets. We report the average token length, relative token reduction, and ROUGE-L scores to illustrate the efficiency-performance trade-off.

\begin{table*}[!t]
\centering
\adjustbox{max width=\textwidth}{
\begin{tabular}{lccc}
\toprule
\textbf{Model} & \textbf{Avg. Token Length} & \textbf{Token Reduction} & \textbf{Avg. ROUGE-L} \\
\midrule
Qwen2.5-32B w/ SC & 1723.83 & - & 50.07 \\
Qwen2.5-32B w/ T$^2$ & 1469.24 & 14.8\% vs. SC & 56.22 \\
\hdashline
QwQ-32B-Preview & 2068.34 & - & 63.38 \\
QwQ-32B w/ T$^2$ & 1547.22 & 25.2\% vs. QwQ & 68.56 \\
\bottomrule
\end{tabular}}
\caption{Efficiency-performance trade-off across seven CQA datasets. Token reduction is calculated relative to the baseline model (SC: Self-Consistency).}
\label{tab:efficiency_performance}
\end{table*}

\subsection{Dataset-specific Efficiency Gains}

\begin{figure*}[!t]
\centering
\includegraphics[width=\textwidth]{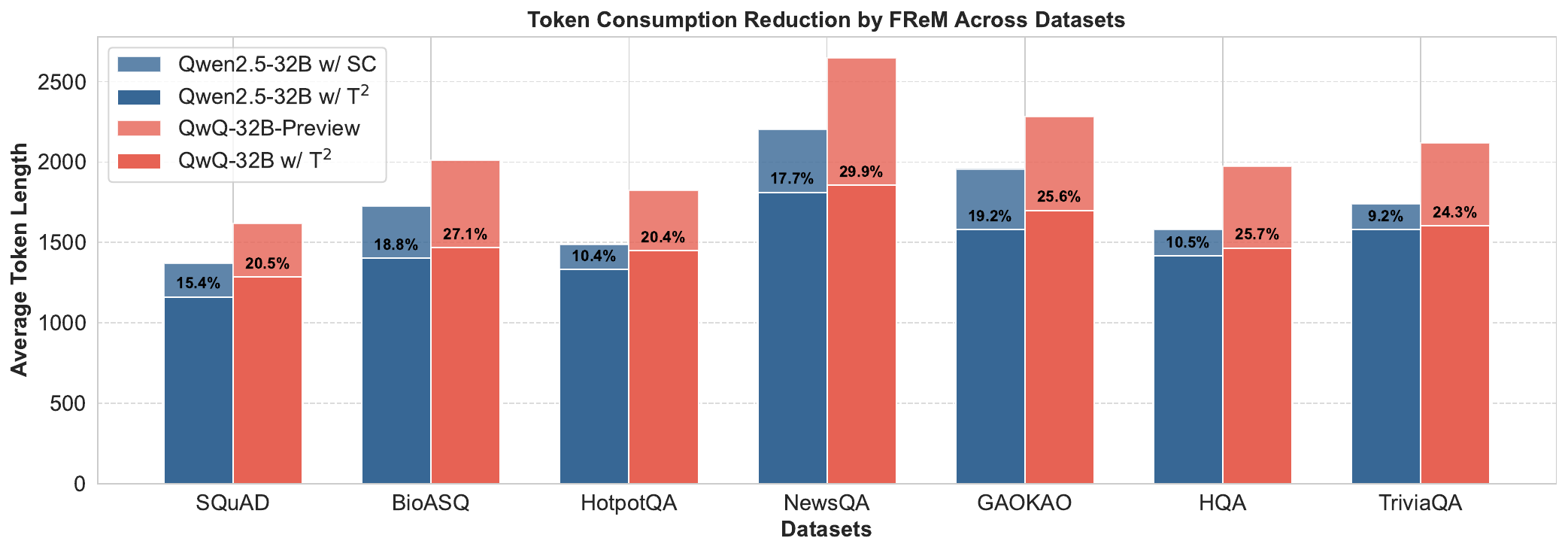}
\caption{Token consumption reduction by T$^2$ across different datasets. The percentage values indicate the relative reduction compared to the baseline models (Qwen2.5-32B w/ SC and QwQ-32B-Preview).}
\label{fig:efficiency_by_dataset}
\end{figure*}

As shown in Figure \ref{fig:efficiency_by_dataset}, the efficiency gains of T$^2$ vary across datasets. The token reduction ranges from 10.5\% to 18.8\% when applied to Qwen2.5-32B (compared to self-consistency), and from 20.6\% to 31.6\% when applied to QwQ-32B (compared to QwQ-32B-Preview). Notably, datasets requiring more complex reasoning (like NewsQA and GAOKAO) show greater efficiency improvements, suggesting that T$^2$ is particularly effective at streamlining the reasoning process for complex questions.

\subsection{Detailed Efficiency-Performance Analysis}

Table \ref{tab:detailed_efficiency} provides a detailed analysis of both token consumption and performance for each dataset and model combination. This comparison highlights how T$^2$ maintains or improves performance while reducing computational costs.

\begin{table*}[!t]
\centering
\adjustbox{max width=\textwidth}{
\begin{tabular}{lcccccccc}
\toprule
\multirow{2}{*}{\textbf{Model}} & \multicolumn{2}{c}{\textbf{SQuAD}} & \multicolumn{2}{c}{\textbf{BioASQ}} & \multicolumn{2}{c}{\textbf{HotpotQA}} & \multicolumn{2}{c}{\textbf{NewsQA}} \\
\cmidrule(lr){2-3} \cmidrule(lr){4-5} \cmidrule(lr){6-7} \cmidrule(lr){8-9}
 & Token & ROUGE-L & Token & ROUGE-L & Token & ROUGE-L & Token & ROUGE-L \\
\midrule
Qwen2.5-32B w/ SC & 1372.18 & 75.31 & 1726.32 & 57.57 & 1485.87 & 56.76 & 2201.65 & 52.27 \\
Qwen2.5-32B w/ T$^2$ & 1161.42 & 81.86 & 1401.52 & 65.02 & 1330.71 & 67.11 & 1812.28 & 61.27 \\
QwQ-32B-Preview & 1617.42 & 86.87 & 2012.33 & 69.02 & 1823.49 & 71.86 & 2648.12 & 63.92 \\
QwQ-32B w/ T$^2$ & 1285.36 & 92.12 & 1467.85 & 75.21 & 1450.75 & 77.61 & 1855.89 & 68.61 \\
\midrule
\multirow{2}{*}{\textbf{Model}} & \multicolumn{2}{c}{\textbf{GAOKAO}} & \multicolumn{2}{c}{\textbf{HQA}} & \multicolumn{2}{c}{\textbf{TriviaQA}} & \multicolumn{2}{c}{\textbf{Average}} \\
\cmidrule(lr){2-3} \cmidrule(lr){4-5} \cmidrule(lr){6-7} \cmidrule(lr){8-9}
 & Token & ROUGE-L & Token & ROUGE-L & Token & ROUGE-L & Token & ROUGE-L \\
\midrule
Qwen2.5-32B w/ SC & 1957.93 & 30.57 & 1580.43 & 37.12 & 1742.41 & 41.92 & 1723.83 & 50.07 \\
Qwen2.5-32B w/ T$^2$ & 1581.14 & 34.06 & 1415.18 & 40.31 & 1582.42 & 43.92 & 1469.24 & 56.22 \\
QwQ-32B-Preview & 2284.80 & 43.23 & 1972.37 & 49.62 & 2119.88 & 59.16 & 2068.34 & 63.38 \\
QwQ-32B w/ T$^2$ & 1699.45 & 47.42 & 1465.68 & 54.71 & 1605.56 & 64.22 & 1547.22 & 68.56 \\
\bottomrule
\end{tabular}}
\caption{Detailed comparison of token consumption and performance (ROUGE-L) across all datasets. Lower token count with higher ROUGE-L indicates better efficiency-performance trade-off.}
\label{tab:detailed_efficiency}
\end{table*}

\subsection{Efficiency Analysis by Question Complexity}

To better understand T$^2$'s efficiency gains, we categorize questions by complexity and analyze token reduction. As shown in Table \ref{tab:complexity_analysis}, T$^2$ achieves greater token reduction for complex questions requiring multi-step reasoning, showcasing its adaptive nature.

\begin{table*}[!t]
\centering
\adjustbox{max width=\textwidth}{
\begin{tabular}{lccc}
\toprule
\textbf{Question Complexity} & \textbf{Qwen2.5 + SC} & \textbf{Qwen2.5 + T$^2$} & \textbf{Token Reduction} \\
\midrule
Simple (1-step) & 1283.45 & 1157.82 & -9.8\% \\
Moderate (2-3 steps) & 1687.31 & 1391.65 & -17.5\% \\
Complex (4+ steps) & 2201.73 & 1758.24 & -20.1\% \\
\midrule
\textbf{Question Complexity} & \textbf{QwQ-32B} & \textbf{QwQ-32B + T$^2$} & \textbf{Token Reduction} \\
\midrule
Simple (1-step) & 1584.21 & 1262.35 & -20.3\% \\
Moderate (2-3 steps) & 2041.57 & 1492.18 & -26.9\% \\
Complex (4+ steps) & 2579.24 & 1887.14 & -26.8\% \\
\bottomrule
\end{tabular}}
\caption{Token consumption analysis by question complexity. T$^2$ achieves greater efficiency gains for more complex questions.}
\label{tab:complexity_analysis}
\end{table*}

\subsection{Time Efficiency}

Beyond token reduction, we also measure the actual inference time across different models and reasoning approaches. Table \ref{tab:time_efficiency} presents the average inference time per question, demonstrating that T$^2$ reduces computational time while maintaining high performance.

\begin{table*}[!t]
\centering
\adjustbox{max width=\textwidth}{
\begin{tabular}{lcc}
\toprule
\textbf{Model} & \textbf{Avg. Inference Time (s)} & \textbf{Time Reduction} \\
\midrule
Qwen2.5-32B w/ SC & 65.31 & - \\
Qwen2.5-32B w/ T$^2$ & 34.52 & -47.1\% \\
\hdashline
QwQ-32B-Preview & 76.74 & - \\
QwQ-32B w/ T$^2$ & 45.03 & -41.3\% \\
\bottomrule
\end{tabular}}
\caption{Average inference time per question across datasets. T$^2$ reduces computational time while maintaining high performance.}
\label{tab:time_efficiency}
\end{table*}

In summary, our comprehensive efficiency analysis demonstrates that T$^2$ reduces token consumption and inference time across diverse CQA datasets while maintaining or improving performance. The efficiency gains are particularly pronounced for complex questions requiring multi-step reasoning, highlighting T$^2$'s ability to adapt its reasoning approach based on question complexity.

\section{Impacts of Similar Examples}

\subsection{Impacts of Question Structure}
\label{apd:ipt_of_q_structure}

Our framework decomposes each question into a \textit{structure} plus \textit{replaceable elements}. We hypothesize that questions with more placeholders benefit more from T$^2$'s selection mechanism, because these questions allow a wider range of possible similar examples. Conversely, simpler questions with fewer placeholders may not need advanced reasoning paths.

We categorize questions into three buckets based on the number of placeholders in $Q$: \textit{Low} (0--1 placeholders), \textit{Medium} (2--3 placeholders), and \textit{High} (4+ placeholders). Table \ref{tab:structure_complexity} shows the performance across these groups for SQuAD and HQA to show impacts on general and domain-specific scenarios.

\begin{table*}[!t]
    \centering
    \begin{tabular}{lcccccc}
        \toprule
        \multirow{2}{*}{\textbf{Group}}& \multicolumn{3}{c}{\textbf{SQuAD}} & \multicolumn{3}{c}{\textbf{HQA}} \\
        \cmidrule(lr){2-4} \cmidrule(lr){5-7}
        & Few-shots & Self-Cconsistency & T$^2$ & Few-shots & Self-Cconsistency & T$^2$ \\
        \midrule
        Low    & 78.5 & 79.1 & \textbf{80.2} & 42.7 & 43.3 & \textbf{44.6} \\
        Medium & 76.4 & 78.2 & \textbf{79.5} & 41.5 & 43.0 & \textbf{45.1} \\
        High   & 75.9 & 78.7 & \textbf{80.1} & 40.2 & 42.9 & \textbf{46.2} \\
        \bottomrule
    \end{tabular}
    \caption{ROUGE-L by question complexity. We compare quick-thinking (Few-shots) and slow-thinking (Self-Consistency), and our T$^2$.}
    \label{tab:structure_complexity}
\end{table*}

As seen in Table \ref{tab:structure_complexity}, questions with more placeholders (High) see the largest gap between T$^2$ and either baseline. This suggests that, for complex questions, enumerating and reusing relevant skill chains is particularly helpful. On simpler questions (Low), T$^2$ still improves performance but by a smaller margin, as fewer placeholders limit the search space for alternative question structures.

\subsection{Impacts of Similar Examples Structure}
\label{apd:ipt_of_d_structure}

We show the ``skeleton'' QA pairs that preserved reasoning structure while replacing all content-specific terms with placeholders in Figure~\ref{fig:sim_example_original} (for original one) and Figure~\ref{fig:sim_example_structure_only} (for structure-only one).

\begin{figure*}[!t]
\begin{tcolorbox}[title=Original QA Pair]
\textbf{Q:} How does the melting point of sodium compare to potassium?

\textbf{A:} I'll approach this systematically:\\
1. First, I'll recall the melting point of sodium, which is 97.8°C.\\
2. Next, I'll recall the melting point of potassium, which is 63.5°C.\\
3. Comparing these values: 97.8°C > 63.5°C\\
4. Therefore, sodium has a higher melting point than potassium.
\end{tcolorbox}
\caption{Our original generated similar example.}
\label{fig:sim_example_original}
\end{figure*}

\begin{figure*}[!t]
\begin{tcolorbox}[title=Structure-Only Version]
\textbf{Q:} How does [PROPERTY] of [ENTITY\_A] compare to [ENTITY\_B]?

\textbf{A:} I'll approach this systematically:\\
1. First, I'll determine the [PROPERTY] of [ENTITY\_A], which is [VALUE\_A].\\
2. Next, I'll determine the [PROPERTY] of [ENTITY\_B], which is [VALUE\_B].\\
3. Comparing these values: [COMPARISON\_OPERATION]\\
4. Therefore, [CONCLUSION\_STATEMENT].
\end{tcolorbox}
\caption{Structure-only version of our generated similar example.}
\label{fig:sim_example_structure_only}
\end{figure*}

\subsection{Impacts of Similar Example Numbers}
\label{apd:ipt_of_d_number}

\begin{figure*}[!t]
    \centering
    \includegraphics[width=\linewidth]{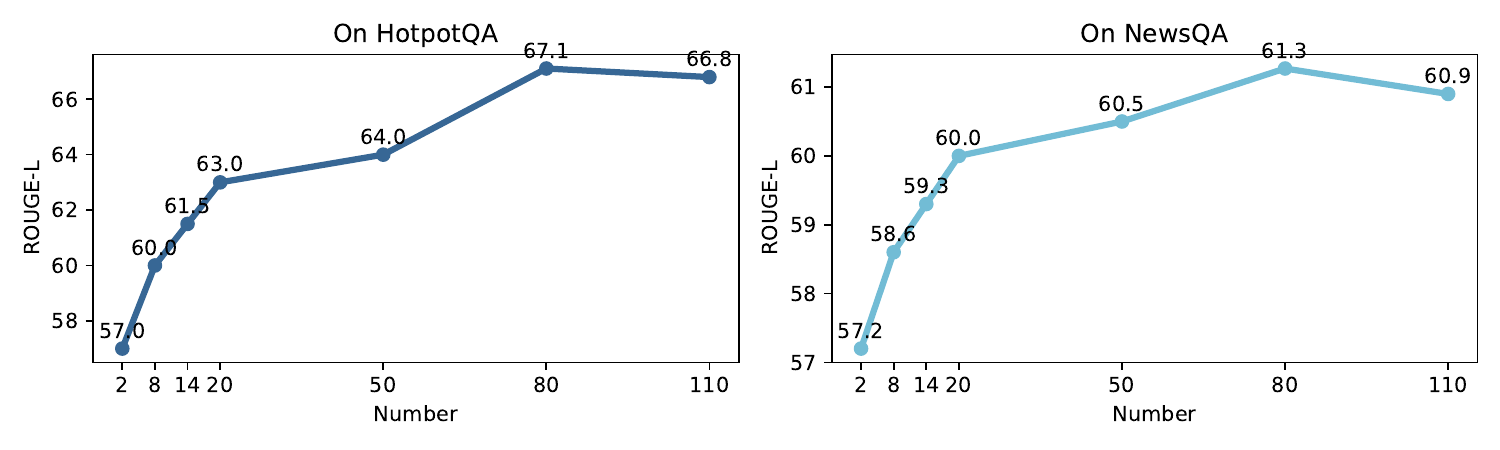}
    \caption{Impact of the number of similar examples on ROUGE-L scores for HotpotQA (left) and NewsQA (right).}
    \label{fig:demo_counts}
\end{figure*}

We vary the size $M=|\Gamma|$. Figure \ref{fig:demo_counts} illustrates the performance on HotpotQA (left) and NewsQA (right) as $M$ increases. We observe an initial boost in ROUGE-L scores before $M=20$, but performance plateaus or slightly decreases beyond a certain point. After increasing examples to $M=80$, the performance rapidly decreases. We conclude that too many examples can introduce irrelevant or redundant paths, making selection harder. In practice, we find that generating a moderate pool is enough to cover essential patterns, especially if the examples are diverse and accurate.

\subsection{Impacts of Example Generation Methods}
\label{apd:ipt_of_d_generate}

\begin{table}[!t]
    \centering
    \adjustbox{max width=\linewidth}{
    \begin{tabular}{lccc}
        \toprule
        \textbf{Method} & \textbf{ROUGE-L} & \textbf{Variation} & \textbf{Noise}\\
        \midrule
        \multicolumn{4}{l}{\textbf{\textit{Qwen2.5 w/ ours}}}\\
        \; Random Fill            & 49.8 & High & Medium \\
        \; Guided Fill            & 52.6 & Low & Low \\
        \; Template Variation   & \textbf{61.3} & High & Low \\
        \bottomrule
    \end{tabular}}
    \caption{Comparing different example construction methods on NewsQA.}
    \label{tab:demo_construction}
\end{table}

Then, we consider how we synthesize reference examples. We experiment with different approaches for filling the placeholders on HotpotQA with qwen2.5-32B:
\begin{itemize}[leftmargin=*,topsep=0pt, partopsep=0pt, itemsep=0pt, parsep=0pt]
    \item \textbf{Random Fill:} Pick random words or entities of the same type (e.g., any \texttt{person}) from a large corpus.
    \item \textbf{Guided Fill:} Use an LLM or curated list to pick semantically relevant or thematically consistent entities for each placeholder.
    \item \textbf{Template Variation:} Generate minor paraphrases or new question stems while retaining the same skill sequence.
\end{itemize}

Table \ref{tab:demo_construction} shows that template variation produces more coherent examples, with 2--4\% gains over purely random fill. This highlights the importance of a well-structured synthetic process: random replacements might yield too many off-topic or contradictory examples, while guided replacements and paraphrasing keep the examples relevant, improving the final answer selection.

\subsection{Impacts of Example Generation Threshold}
\label{apd:ipt_of_d_threshold}

\begin{figure*}[!t]
    \centering
    \includegraphics[width=\linewidth]{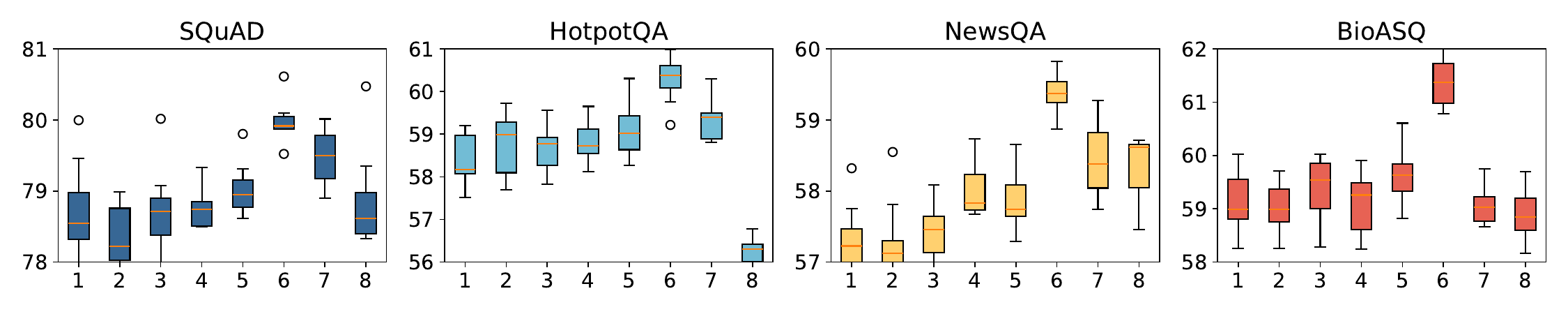}
    \caption{Impact of the question synthesis scope.}
    \label{fig:impact_of_delta}
\end{figure*}

We analyze the impact of varying the threshold $\delta$ on the synthesis quality of the generated questions. The threshold \(\delta\) controls how similar the synthesized questions \(Q_{\mathrm{syn}}^i\) are to the original question \(Q\), by using a helper language model to assess their alignment (in Sec.\ref{sec:demo-selection}). Figure \ref{fig:impact_of_delta} shows finding a trade-off between question similarity and generalization is much more important. As \(\delta\) increases, the similarity to the original question improves but at the cost of generalization. Conversely, when \(\delta\) is lowered, the model generalizes better but the quality of the synthesized questions decreases.

\subsection{Impacts of Examples Domain Bias and Structural Bias}
\label{apd:ipt_of_d_domain}

\begin{table}[!t]
\centering
\small
\begin{tabular}{lcc}
\toprule
\textbf{Model} & \textbf{HotpotQA} & \textbf{HQA} \\
\midrule
Qwen2.5+SC & 56.76 & 37.12 \\
Qwen2.5+T$^2$ & 67.11 & 40.31 \\
Qwen2.5+mis domain & 65.96 & 39.85 \\
Qwen2.5+structure only & 63.96 & 38.85 \\
\hdashline
QwQ & 71.86 & 49.62 \\
QwQ+T$^2$ & 77.61 & 54.71 \\
QwQ+mis domain & 77.03 & 54.26 \\
QwQ+structure only & 74.03 & 53.66 \\
\bottomrule
\end{tabular}
\caption{Performance on mis-domain and structure-only models. ROUGE-L is the reported performance metric.}
\label{tab:demo_quality}
\end{table}

In addition, we investigate the effects of domain and structural biases in similar examples. Specifically, we assess how varying the domain of the similar examples influences model performance. As shown in the Table~\ref{tab:demo_quality} (``+mis domain''), transitioning from a general domain to a historical one results in improved performance compared to using self-consistency alone. Furthermore, we evaluate the impact of removing key information from the similar examples, leaving only the reasoning structure. Table~\ref{tab:demo_quality} (``+structure only'') demonstrates that even when only the examples' structure\footnote{We show the example structure in Appendix \ref{apd:ipt_of_d_structure}.} is provided, the model can still generate appropriate responses, highlighting the effectiveness of structural guidance.

\subsection{Impacts of Similar Examples Quality}
\label{apd:ipt_of_d_quality}

To evaluate the quality of similar examples generated by our framework, we conducted a comprehensive human evaluation study. We randomly selected 1000 query-reference pairs from the HotpotQA dataset and recruited three Ph.D. students specializing in NLP to assess the quality of synthetic references. The evaluation was conducted blind, with evaluators unaware of which model generated each reference.

\paragraph{Evaluation Dimensions.} References were rated on a scale of 1-10 across four key dimensions:
\begin{itemize}[leftmargin=*, itemsep=0pt,parsep=0pt,topsep=0pt,partopsep=0pt]
    \item \textbf{Accuracy:} Factual correctness and absence of hallucinations or contradictions
    \item \textbf{Relevance:} Degree to which the reference addresses the specific query requirements
    \item \textbf{Completeness:} Thoroughness in covering all necessary information and reasoning steps
    \item \textbf{Coherence:} Logical structure, clarity of expression, and overall readability
\end{itemize}

\paragraph{Model Comparison.} We evaluated synthetic references generated by two foundation models: Qwen2.5-32B-Instrucut and QwQ-32B-Preview, both with our framework. Table~\ref{tab:human_eval} presents the average scores across all evaluators and samples.

\begin{table*}[!t]
\centering
\begin{tabular}{lccccc}
\toprule
\textbf{Models w/ T$^2$} & \textbf{Accuracy} & \textbf{Relevance} & \textbf{Completeness} & \textbf{Coherence} & \textbf{Overall} \\
\midrule
Qwen2.5-32B-Instrucut & 8.8 & 8.6 & 8.1 & 8.1 & 8.4 \\
QwQ-32B-Preview & 8.4 & 8.3 & 7.9 & 8.2 & 8.4 \\
\bottomrule
\end{tabular}
\caption{Human evaluation scores for synthetic references generated by different models (scale: 1-10).}
\label{tab:human_eval}
\end{table*}

Results show both models produced high-quality references. The highest scores were observed in the Relevance category, indicating that references effectively addressed the specific queries. The evaluation exhibited strong inter-annotator agreement with a Fleiss' kappa coefficient of 0.79, indicating substantial agreement among the three evaluators. This suggests the evaluation results are reliable and consistent across different human judges.

\section{Detailed Case Studies}
\label{apd:case_studies}

Figure \ref{fig:hotpotqa_example} and \ref{fig:squad_case_single_step} show the two different cases from HotPotQA and SQuAD. The two case studies illustrate distinct reasoning strategies for question answering. In the HotpotQA case, the task requires multi-step reasoning by integrating evidence from multiple documents. A response based solely on pattern matching might output wrong ``London'' and an overthinking answer may include unnecessary details before arriving at the correct conclusion, the best approach is a concise, step-by-step explanation that clearly connects the film to John Lennon and his documented birthplace.

In contrast, the SQuAD case involves single-step reasoning where the answer is directly available from the text. The document states that the Eiffel Tower was constructed in 1889, so the correct answer is deduced by simply extracting this fact. Although a pattern matching response may correctly state ``1889'' and an overcomplicated explanation might unnecessarily discuss other numerical details, the ideal answer succinctly identifies the key fact and deduces that the construction year is 1889 using clear, deductive reasoning.

\begin{figure*}[!t]
    \small
    \begin{tcolorbox}[title=\# HotpotQA Case Example with Multihop Thinking, fonttitle=\bfseries, colframe=black, colback=white, sharp corners, boxrule=0.5mm, colbacktitle=gray!20, coltitle=black, enhanced]
        \textbf{Document 1:}\\
        John Lennon, the iconic musician and member of The Beatles, was born in Liverpool, England.\\        
        \textbf{Document 2:}\\
        \textit{Nowhere Boy} is a film that explores the early life of John Lennon, highlighting his formative years and influences.\\
        \textbf{Question:}\\
        In what city was the subject of the film \textit{Nowhere Boy} born?

        \vspace{1em}
        
        \textbf{Proper Reasoning Chain:}\\
        1. \textbf{Identify the Subject:} Recognize that \textit{Nowhere Boy} is about John Lennon. (Deductive)\\
        2. \textbf{Extract Key Fact:} From Document 1, note that John Lennon was born in Liverpool, England. (Deductive)\\
        3. \textbf{Conclude:} Therefore, the subject of the film was born in Liverpool. (Decompositional)

        \vspace{1em}
        
        \textcolor{red}{\textbf{1) Quick Thinking Model's Wrong Answer:}}\\
        \textcolor{red}{\textit{Answer:} The subject of \textit{Nowhere Boy} was born in London.}\\
        
        \textcolor{orange}{\textbf{2) Slow Thinking Model's Overthinking Answer:}}\\
        \textcolor{orange}{\textit{Answer:} After a lengthy analysis of various biographical details concerning John Lennon, including his early childhood, multiple hints in his music and interviews, and a discussion of several English cities mentioned in relation to his life, one might eventually conclude that the subject of \textit{Nowhere Boy} was born in Liverpool.}\\
        
        \textcolor{mygreen}{\textbf{3) Model with Our T$^2$'s Correct Answer:}}\\
        \textcolor{mygreen}{\textit{Answer:} Since \textit{Nowhere Boy} is a film about John Lennon (Document 2) and Document 1 confirms that John Lennon was born in Liverpool, we deduce that the subject of the film was born in Liverpool.}
    \end{tcolorbox}
    \caption{HotpotQA Case Example with Multiple Answer Versions and Explicit Reasoning Chain.}
    \label{fig:hotpotqa_example}
\end{figure*}

\begin{figure*}[!t]
    \small
    \begin{tcolorbox}[title=\# SQuAD Case Example with Single-Step Thinking,fonttitle=\bfseries,colframe=black,colback=white,sharp corners,boxrule=0.5mm,colbacktitle=gray!20,coltitle=black,enhanced]
        \textbf{Document:}\\
        The Eiffel Tower is a wrought-iron lattice tower located on the Champ de Mars in Paris, France. It was constructed in 1889 as the entrance to the 1889 World's Fair. The tower stands approximately 324 meters tall and is one of the most recognized structures in the world.\\
        \textbf{Question:}\\
        In what year was the Eiffel Tower constructed?\\

        \vspace{1em}
        
        \textbf{Proper Reasoning Chain (using Deductive):}\\
        1) \textbf{Deduction:} Based on clear statement, we deduce that the construction year is 1889 (Deductive).
        
        \vspace{1em}
        
        \textbf{1) Quicking Thinking Model's Correct Answer \textcolor{orange}{via Pattern Matching}:}\\
        \textit{Answer:} 1889.\\
        
        \textbf{2) Slow Thinking Model's Correct Answer \textcolor{orange}{but Overthinking}:}\\
        \textit{Answer:} After reading the passage, I noticed several numbers such as 324 (the tower's height) and 1889. While I considered whether the height might play a role in determining the construction year, I ultimately realized that the passage explicitly states ``constructed in 1889.'' Thus, despite the extra details, the correct construction year is 1889.\\
        
        \textbf{3) Model with Our T$^2$'s Correct Answer \textcolor{mygreen}{using the Specific Reasoning Chain}:}\\
        \textit{Answer:} Since the document clearly states that the Eiffel Tower was constructed in 1889, we deduce—by directly extracting and applying this fact—that the construction year is 1889.
    \end{tcolorbox}
    \caption{SQuAD Case Example with Single-Step Thinking and Multiple Answer Versions.}
    \label{fig:squad_case_single_step}
\end{figure*}

\end{document}